\PassOptionsToPackage{table}{xcolor}
\documentclass[sigconf]{aamas}
\usepackage{algorithm}
\usepackage{algpseudocode}
\usepackage{balance}
\usepackage{soul}
\usepackage{xcolor}
\usepackage{enumitem}
\sethlcolor{yellow}
\usepackage{comment}
\soulregister\cite7
\usepackage{amsmath,amsfonts}
\usepackage{graphicx,booktabs,multirow}
\usepackage{nicefrac,microtype,soul,etoc}
\setcopyright{ifaamas}
\acmConference[AAMAS '26]{Proc.\@ of the 25th International Conference
on Autonomous Agents and Multiagent Systems (AAMAS 2026)}{May 25 -- 29, 2026}
{Paphos, Cyprus}{C.~Amato, L.~Dennis, V.~Mascardi, J.~Thangarajah (eds.)}
\copyrightyear{2026}
\author{Taozheng Zhu}
\affiliation{%
  \institution{University of Notre Dame}
  \city{Notre Dame}
  \state{Indiana}
  \country{USA}
}
\email{tzhu2@nd.edu}

\author{Shixiong Jiang}
\affiliation{%
  \institution{University of Notre Dame}
  \city{Notre Dame}
  \state{Indiana}
  \country{USA}
}
\email{sjiang5@nd.edu}

\author{Fanxin Kong}
\authornote{Corresponding author.}
\affiliation{%
  \institution{University of Notre Dame}
  \city{Notre Dame}
  \state{Indiana}
  \country{USA}
}
\email{fkong@nd.edu}
\acmYear{2026}
\acmDOI{}
\acmPrice{}
\acmISBN{}
\acmSubmissionID{}

\title{Vulnerability Analysis for Safe Multi-Agent Reinforcement Learning}
\begin{abstract}
Safe Multi-Agent Reinforcement Learning (SMARL) aims to learn cooperative control policies under prescribed safety constraints. Although existing SMARL methods focus on enforcing safety during policy optimization, their vulnerability remains largely unexplored. This paper investigates inference-time observation attacks against SMARL, where an adversary perturbs the observations of a subset of agents, thereby altering the actions produced by their policies. We propose MinMax Attack, an attack framework that jointly selects adversarial target actions for compromised agents while accounting for the Cooperative Resistance of benign agents. We further derive a bound on the induced safety cost and characterize its relationship with the Compromised-Agent Budget and the Perturbation Budget. Experiments across multiple safety-critical multi-agent benchmarks and representative state-of-the-art SMARL algorithms show that MinMax Attack induces substantially higher safety costs than existing attack baselines while preserving task reward, thereby reducing its detectability and achieving both attack effectiveness and stealthiness.
\end{abstract}

\begin{document}
\pagestyle{fancy}
\fancyhead{}
\maketitle

\section{Introduction}
As Multi-Agent Reinforcement Learning (MARL) is increasingly deployed in cooperative real-world systems, it provides a learning framework for coordinated control problems, such as path planning for robot and drone swarms~\cite{53_qin2021learning}, distributed voltage control in power grids~\cite{54_wang2021multi}, and large-scale traffic signal control~\cite{55_liu2023gplight}. Ensuring the safe operation of such multi-agent systems is therefore critical. Safe Multi-Agent Reinforcement Learning (SMARL) studies how multiple agents can optimize cooperative policies while satisfying prescribed safety constraints. Intuitively,  safety constraints should make the system more robust to attacks. Our study reveals a different possibility: the rules introduced by SMARL to enforce safety can also create an attack surface for adversaries. Since agents trust and coordinate with other agents to remain within shared safety boundaries, their behaviors become more predictable, making them more vulnerable to targeted attacks.

Prior studies have shown that MARL systems remain vulnerable despite substantial algorithmic progress \cite{36_he2021secure, 37_bijani2014review}. Recent attacks further expose this vulnerability. For example, SAJA jointly perturbs observations and actions at inference time to degrade multi-agent performance \cite{28_guo2025saja}. However, existing MARL attacks primarily target cumulative reward degradation, without explicitly optimizing safety constraint violations \cite{72_lin2020robustness, 73_hu2022sparse}. Directly transferring safety-oriented attacks from single-agent RL to SMARL can also be suboptimal, as their effects are mediated by inter-agent coordination and may be offset by benign agents' responses \cite{74_li2025attacking, 75_lee2025wolfpack}. Thus, the adversarial vulnerability of SMARL remains largely unexplored.

To address this gap, we propose MinMax Attack, an inference-time observation attack that explicitly targets safety violations in SMARL. It coordinates observation perturbations across multiple compromised agents while accounting for the Cooperative Resistance of benign agents under coupled safety constraints. Extending prior observations on benign-agent responses in MARL \cite{74_li2025attacking, 75_lee2025wolfpack}, we formalize Cooperative Resistance in SMARL as a joint action response of
benign agents that locally counteracts the safety cost increase induced by compromised agents. By accounting for Cooperative Resistance, MinMax Attack maximizes safety violations against the strongest benign response.

To the best of our knowledge, this work is the first to study the vulnerability of SMARL under adversarial attacks. The primary contributions of this work are three-fold:
\begin{enumerate}[topsep=0pt,itemsep=0pt,leftmargin=*]
 \item \textbf{An adversarial attack framework for SMARL.} We introduce MinMax Attack, an effective inference-time observation attack that injects jointly targeted perturbations into the local observations of a small subset of agents while accounting for Cooperative Resistance to strengthen attack effectiveness. Across representative SMARL environments and algorithms, MinMax Attack achieves the highest safety cost while maintaining nominal task reward among SOTA observation attack baselines. 
    
    \item \textbf{Theoretical bounds on safety violations.} 
     We prove that the safety cost increase induced by observation attacks admits an upper bound. The bound scales as $\mathcal{O}(m^2)$ with the number of compromised agents $m$ and linearly with the Perturbation Budget, suggesting that compromising more agents can potentially amplify safety cost increase more substantially than increasing the perturbation magnitude.

    \item \textbf{Effectiveness and Stealthiness Validation.} Through comprehensive experiments across multiple representative SMARL environments and algorithms, we show that the proposed MinMax Attack induces severe safety violations while preserving nominal task performance, demonstrating both its effectiveness and stealthiness.  
\end{enumerate}
   

\section{Related Work}
This section reviews two major lines of related work: Safe Multi-Agent Reinforcement Learning (SMARL) and adversarial attacks on Safe Reinforcement Learning (SRL) and MARL.
\subsection{SMARL}

MARL aims to learn cooperative policies when each agent can only observe a local part of the environment. Centralized training with decentralized execution has therefore become a foundational paradigm for MARL \cite{31_zhang2021multi,32_gronauer2022multi,33_zhu2024survey}. Prior work improves this paradigm by using shared critics to guide coordination \cite{23_lowe2017multi,24_peng2021facmac}, estimating the individual contribution of each agent to team performance \cite{30_foerster2018counterfactual}, and decomposing a shared team objective into tractable individual learning problems \cite{22_rashid2020monotonic}.

SMARL builds on the safety mechanisms developed in single-agent safe reinforcement learning. Single-agent safe RL studies how to maximize task rewards while satisfying strict safety constraints \cite{25_gu2024review,34_garcia2015comprehensive}. Many methods enforce safety constraints through Lagrange multipliers and gradient based policy updates \cite{3_zhang2020first,8_kim2024conflict,9_so2024solving}. Recent methods further improve standard Lagrangian updates by incorporating proportional integral derivative control \cite{7_stooke2020responsive,40_johnson2005pid} and probabilistic inference frameworks \cite{4_liu2022constrained,5_yao2023constraint}. Other techniques address multiple safety constraints and reduce out of distribution extrapolation errors through gradient shaping \cite{1_yao2024gradient} and variational autoencoders \cite{6_xu2022constraints,10_guo2025constraint}.

Based on these safety mechanisms, SMARL models cooperative multi-agent systems as constrained Markov games. Its goal is to maximize team reward while preventing agents from crossing critical safety boundaries \cite{15_zhou2024multi,26_gu2023safe}. Recent SMARL methods further improve scalability by restricting information sharing to direct neighbors within $k$-hop neighborhoods, while still providing mathematical guarantees for global safety \cite{20_zhang2024scalable}. These advances enable SMARL to support safety critical applications such as autonomous vehicle fleets and unmanned aerial vehicle swarms \cite{38_dinneweth2022multi,39_azar2021drone}.

\subsection{Adversarial Attacks on MARL}
Deep neural networks are vulnerable to small, often imperceptible adversarial input perturbations \cite{42_goodfellow2014explaining}. Goodfellow et al. introduced the fast gradient sign method (FGSM), which generates adversarial examples through a single gradient-based perturbation step. Madry et al. further strengthened first-order attacks through iterative projected gradient descent \cite{41_madry2017towards}.

Deep reinforcement learning inherits the adversarial vulnerability of deep neural networks. Huang et al. demonstrate that gradient based perturbations on state observations can significantly degrade policy performance and cumulative reward \cite{44_fhuang2017adversarial}. Recent work on SRL shows that adversaries can exploit critic information to guide observation perturbations toward safety violations, and further provides a theoretical explanation for how the reward--safety trade-off can cause task reward and safety cost to increase simultaneously, thereby improving stealthiness~\cite{45_liu2022robustness}. Adversarial attacks have also been extended to MARL. During training, attackers can poison online MARL by injecting misleading information into the learning process, thereby degrading the learned policy performance \cite{29_liu2023efficient}. During inference, attackers can manipulate the observations and actions of victim agents to reduce team reward \cite{28_guo2025saja}.

Recent MARL attacks increasingly exploit inter-agent dependencies to amplify adversarial effects. Training-time backdoor attacks leverage team interactions to propagate the influence of compromised agents through benign teammates \cite{65_chen2022marnet,66_fang2026blast,67_chen2025action}. Similar mechanisms have also been explored at inference time. AMI exploits agent-wise influence to induce harmful responses, while Wolfpack targets assisting agents that respond to an initially attacked agent \cite{74_li2025attacking,75_lee2025wolfpack}. These studies show that benign-agent responses can substantially affect attack outcomes in general MARL, primarily through task reward. Unlike observation-only attacks that perturb only agents' observations, AMI requires direct control of a malicious agent, while Wolfpack assumes attack access to subsequently identified assisting agents.

However, attacks that target the safety of SMARL systems remain barely studied. Our work addresses this gap by studying the inference-time observation attack against SMARL, where the adversary aims to increase safety cost rather than merely degrade task reward.

\section{Preliminaries}
\subsection{SMARL Formalisms}
Following representative SMARL formulations \cite{26_gu2023safe,46_liu2026safe}, we model SMARL as a fully cooperative constrained Markov game
\[
\mathcal{G}
=
\big\langle
\mathcal{N},\mathcal{S},\mathcal{A},p,\rho_0,\gamma,R,\mathcal{C},c
\big\rangle ,
\]
where $\mathcal{N}=\{1,\ldots,n\}$ is the agent set and $\mathcal{S}$ is the global state space.
The joint action space is $\mathcal{A}=\prod_{i=1}^{n}\mathcal{A}^i,$
where $\mathcal{A}^i$ denotes the action space of agent $i$. A joint action is denoted by
$\boldsymbol{a}=(a^1,\ldots,a^n)\in\mathcal{A}$, where $a^i\in\mathcal{A}^i$ is the action of agent $i$. 
The system evolves according to the transition kernel $p:\mathcal{S}\times\mathcal{A}\times\mathcal{S}\to[0,1],$
with initial state distribution $\rho_0$. The team reward function is $R:\mathcal{S}\times\mathcal{A}\to\mathbb{R}$. 
Safety is specified by a set of agent-wise cost functions
$\mathcal{C}=\{C^i\}_{i\in\mathcal{N}}$, where
$C^i:\mathcal{S}\times\mathcal{A}\to\mathbb{R}_{\ge 0}$ denotes the safety cost of agent $i$, and $c=\{c^i\}_{i\in\mathcal{N}}$ denotes the corresponding safety thresholds.

At timestep $t$, the environment is in state $s_t\in\mathcal{S}$.
Each agent $i$ receives a local policy input $s_t^i$ derived from
the global state and samples an action
$a_t^i\sim\pi^i(\cdot\mid s_t^i)$.
The resulting joint action is
$\mathbf{a}_t=(a_t^1,\ldots,a_t^n)$, and the factorized joint policy is
\[
\pi(\mathbf{a}_t\mid s_t)
=
\prod_{i=1}^{n}\pi^i(a_t^i\mid s_t^i).
\]
The system receives reward $R(s_t,\mathbf{a}_t)$, while agent $i$
incurs safety cost $C^i(s_t,\mathbf{a}_t)$, and then transitions to
$s_{t+1}\sim p(\cdot\mid s_t,\mathbf{a}_t)$.

The SMARL objective is to maximize the expected reward defined as: 
\begin{equation}
J_R(\pi)
=
\mathbb{E}_{\substack{
s_0\sim\rho_0,\;
\mathbf{a}_t\sim\pi(\cdot\mid s_t),\\
s_{t+1}\sim p(\cdot\mid s_t,\mathbf{a}_t)
}}
\left[
\sum_{t=0}^{\infty}
\gamma^t R(s_t,\mathbf{a}_t)
\right].
\end{equation}

the $\gamma\in[0,1)$ is discount factor. While satisfying the safety constraints of every agent $i \in \mathcal{N}$:

\begin{equation}
J_C^i(\pi)
=
\mathbb{E}_{\substack{
s_0\sim\rho_0,\;
\mathbf{a}_t\sim\pi(\cdot\mid s_t),\\
s_{t+1}\sim p(\cdot\mid s_t,\mathbf{a}_t)
}}
\left[
\sum_{t=0}^{\infty}
\gamma^t C^i(s_t,\mathbf{a}_t)
\right]
\le c^i.
\end{equation}

We define the team safety cost as
$J_C(\boldsymbol{\pi})=\sum_{i\in\mathcal{N}}J_C^i(\boldsymbol{\pi})$,
which measures system-level safety degradation. Unless otherwise specified, all safety cost considered in the following refer to this team-level quantity.
Because each $C^i(s,\boldsymbol{a})$ depends on the joint action, the resulting safety effect may exhibit non-separable interactions across agents.

\subsection{Threat Model}
In this work, we formulate a malicious adversary targeting the inference phase of SMARL. We characterize the threat model based on the adversary's knowledge, capability, and objective.

\textbf{Adversary Knowledge.}
Following the widely adopted white-box threat model in adversarial RL and MARL, we assume that the adversary has access to the global state observation and the specific algorithms used by the victim agents, as defined in prior work~\cite{44_fhuang2017adversarial,29_liu2023efficient, 56_gleave2019adversarial}. White-box attacks also serve as a principled security-auditing tool for exposing vulnerabilities that weaker evaluations may miss~\cite{60_carlini2019evaluating, 61_tramer2020adaptive}. This assumption allows us to evaluate the system's safety boundaries under worst-case scenarios.  

\textbf{Adversary Capability.}
We focus on inference-time observation attacks, a widely adopted setting in adversarial RL and MARL~\cite{44_fhuang2017adversarial,48_zhang2020robust,45_liu2022robustness}, where the adversary intercepts sensor data before policy execution and perturbs only the observations of compromised agents. Following prior MARL attack settings~\cite{28_guo2025saja,57_sun2022certifiably}, the adversary can compromise only a budgeted subset of agents. The compromised agents are randomly chosen to prevent arbitrary victim selection from amplifying the attack effect. Let $\mathcal{M}\subseteq\mathcal{N}$ denote the compromised agent subset with Compromised-Agent Budget $|\mathcal{M}|=m$, and let $\mathcal{U}=\mathcal{N}\setminus\mathcal{M}$ denote the remaining benign agents. For each compromised agent $i\in\mathcal{M}$, the adversary applies a bounded perturbation only to its local observation and generates
\[
\hat{s}_t^i=s_t^i+\delta_t^i,
\qquad
\|\delta_t^i\|_p\le\epsilon,
\]
where $\delta_t^i$ is the observation perturbation and $\epsilon$ is the Perturbation Budget. Following existing observation attack settings in adversarial RL~\cite{45_liu2022robustness,62_pattanaik2017robust,63_oikarinen2021robust}, this $\ell_p$ constraint prevents arbitrary observation replacement and restricts the adversary to locally bounded perturbations, such as sensor errors or state-estimation uncertainty. For each benign agent $j\in\mathcal{U}$, the policy input remains unchanged.

Under the attack, the executed joint policy becomes
\[
\tilde{\boldsymbol{\pi}}_{\delta}(\boldsymbol{a}_t\mid \boldsymbol{s}_t)
=
\prod_{i\in\mathcal{M}}
\pi^i(a_t^i\mid \hat{s}_t^i)
\prod_{j\in\mathcal{U}}
\pi^j(a_t^j\mid s_t^j).
\]

Compared with MARL attacks requiring stronger intervention, our setting uses only bounded observation perturbations. AMI requires direct control of a malicious agent, Wolfpack assumes access to assisting agents, and SAJA manipulates both observations and actions~\cite{74_li2025attacking,75_lee2025wolfpack,28_guo2025saja}. This observation attack's setting better reflects sensor-level interference at deployment.

\textbf{Adversary Objective.}
The adversary aims to induce safety violations, quantified by an increase in safety cost, while remaining stealthy to reward-based monitoring. Prior studies have shown that adversarial behavior
can be detected through abnormal changes in reward-related signals,
\cite{64_havens2018online,76_mo2024security}. Following prior safe RL attacks,
we therefore characterize attack effectiveness by safety cost increase and
stealthiness by reward preservation \cite{45_liu2022robustness}.

Let
$\Delta_C
=
J_C(\tilde{\boldsymbol{\pi}}_{\delta})
-
J_C(\boldsymbol{\pi})$ denotes the attack-induced increase in team safety cost, and let $\Delta_R
=
J_R(\boldsymbol{\pi})
-
J_R(\tilde{\boldsymbol{\pi}}_{\delta})$ denotes the corresponding task reward degradation.

\begin{definition}[(Attack) Effectiveness]
\label{def:effectiveness}
An attack is effective if it increases the safety cost, i.e.,
$\Delta_C>0$.
\end{definition}

\begin{definition}[(Reward) Stealthiness]
\label{def:stealthiness}
An attack is stealthy if its task reward degradation satisfies
$\Delta_R\le\eta_R$, where $\eta_R\ge0$ denotes the tolerated reward
degradation under reward-based monitoring.
\end{definition}

Accordingly, the adversary maximizes safety cost increase subject to the two budgets and stealthiness requirement
\begin{equation}
\label{eq:attack_objective}
\begin{aligned}
\max_{\{\delta_t^i\}_{i\in\mathcal{M},t\ge0}}\quad
&J_C(\tilde{\boldsymbol{\pi}}_{\delta})\\
\mathrm{s.t.}\quad
&\|\delta_t^i\|_p\le\epsilon,
\quad i\in\mathcal{M},\ t\ge0,\\
&J_R(\tilde{\boldsymbol{\pi}}_{\delta})
\ge
J_R(\boldsymbol{\pi})-\eta_R.
\end{aligned}
\end{equation}

\subsection{Multi-Agent Safety Interactions}
\label{sec:safety_interactions}
The SMARL setting exhibits two safety-relevant interactions: coupling among compromised agents and compensatory responses from benign agents.

Let $\mathbf{a}_{\mathcal M}$ and $\mathbf{a}_{\mathcal U}$
denote the joint actions of the compromised and benign agents,
respectively, and let $Q_c^{\boldsymbol{\pi}}(s,\boldsymbol{a})$ denote the safety cost
critic under the clean policy $\boldsymbol{\pi}$. Let $\boldsymbol{a}_{\mathcal M}^0$ denote the nominal actions of the
compromised agents. For fixed $\boldsymbol{a}_{\mathcal U}$, we define the
joint safety cost change as
\begin{equation}
\Delta_{\mathrm{joint}}
(s;\boldsymbol{a}_{\mathcal M},\boldsymbol{a}_{\mathcal U})
=
Q_c^{\boldsymbol{\pi}}
(s,\boldsymbol{a}_{\mathcal M},\boldsymbol{a}_{\mathcal U})
-
Q_c^{\boldsymbol{\pi}}
(s,\boldsymbol{a}_{\mathcal M}^0,\boldsymbol{a}_{\mathcal U}),
\end{equation}
and the corresponding single-agent change for each $i\in\mathcal M$ as
\begin{equation}
\Delta_i(s;a^i,\boldsymbol{a}_{\mathcal U})
=
Q_c^{\boldsymbol{\pi}}
(s,a^i,\boldsymbol{a}_{\mathcal M\setminus\{i\}}^0,
\boldsymbol{a}_{\mathcal U})
-
Q_c^{\boldsymbol{\pi}}
(s,\boldsymbol{a}_{\mathcal M}^0,\boldsymbol{a}_{\mathcal U}).
\end{equation}

\begin{definition}[Non-separable safety coupling]
\label{def:nssc}
The non-separable safety coupling (NSSC) among compromised agents is
\begin{equation}
\Delta_{\mathrm{NSSC}}
=
\Delta_{\mathrm{joint}}
-
\sum_{i\in\mathcal M}\Delta_i .
\end{equation}
\end{definition}

Thus, $\Delta_{\mathrm{NSSC}}\neq0$ indicates that simultaneously changing
multiple compromised agents produces a safety effect that is not captured
by the sum of their single-agent effects.

Let $\boldsymbol{a}_{\mathcal M}^{\mathrm{ind}}$ denote the joint action
formed by independently optimized compromised-agent actions. For fixed
$(s,\boldsymbol{a}_{\mathcal U})$, let
$J_{\mathrm{ind}}^\star$ and $J_{\mathrm{joint}}^\star$ denote the optimal
independent and joint attack values, respectively.

\begin{proposition}[]
\label{prop:joint_attack}
The optimal joint attack value satisfies
\begin{equation}
J_{\mathrm{joint}}^\star(s,\boldsymbol{a}_{\mathcal U})
\ge
J_{\mathrm{ind}}^\star(s,\boldsymbol{a}_{\mathcal U})
+
\Delta_{\mathrm{NSSC}}
\left(
s;\boldsymbol{a}_{\mathcal M}^{\mathrm{ind}},
\boldsymbol{a}_{\mathcal U}
\right).
\label{eq:joint_dominance}
\end{equation}
\end{proposition}

When $\Delta_{\mathrm{NSSC}}
(s;\boldsymbol{a}_{\mathcal M}^{\mathrm{ind}},
\boldsymbol{a}_{\mathcal U})\ge0$,
joint optimization is no worse than independent optimization. This result
motivates Jointly Optimized Multi-Victim Action Selection (JOMAS), which
jointly optimizes the target actions of compromised agents. The proof is provided in Appendix~B.

Motivated by the safety-constrained formulation of SMARL, we next consider the response of benign agents. Let
$\boldsymbol{a}^0=
(\boldsymbol{a}_{\mathcal M}^0,\boldsymbol{a}_{\mathcal U}^0)$
denote a nominal joint action at a safety boundary at $s$
$Q_c^{\boldsymbol{\pi}}(s,\boldsymbol{a}^0)=\bar c_s$, where equality indicates that the safety constraint is active. we consider non-degenerate boundary points where their actions can locally affect the safety cost. Accordingly, we assume that
$Q_c^{\boldsymbol{\pi}}$ is differentiable at $(s,\boldsymbol{a}^0)$ and
$\nabla_{\boldsymbol{a}_{\mathcal U}}
Q_c^{\boldsymbol{\pi}}(s,\boldsymbol{a}^0)\neq0$.

A compromised agent deviation $\boldsymbol{d}_{\mathcal M}$ is
cost-increasing if
\begin{equation}
\nabla_{\boldsymbol{a}_{\mathcal M}}
Q_c^{\boldsymbol{\pi}}(s,\boldsymbol{a}^0)^\top
\boldsymbol{d}_{\mathcal M}>0.
\end{equation}
 
\begin{definition}[Cooperative Resistance]
\label{def:cooperative_resistance}
Cooperative Resistance is a benign-agent response
$\boldsymbol{\nu}_{\mathcal U}$ that reduces the safety cost,
i.e.,
\begin{equation}
\nabla_{\boldsymbol{a}_{\mathcal U}}
Q_c^{\boldsymbol{\pi}}(s,\boldsymbol{a}^0)^\top
\boldsymbol{\nu}_{\mathcal U}<0.
\end{equation}
\end{definition}

\begin{lemma}
\label{lem:cooperative_resistance}
For any cost-increasing deviation $\boldsymbol{d}_{\mathcal M}$, there
exists a Cooperative Resistance $\boldsymbol{\nu}_{\mathcal U}$ such that

\begin{equation}
\nabla_{\boldsymbol{a}_{\mathcal U}}
Q_c^{\boldsymbol{\pi}}(s,\boldsymbol{a}^0)^\top
\boldsymbol{\nu}_{\mathcal U}
=
-
\nabla_{\boldsymbol{a}_{\mathcal M}}
Q_c^{\boldsymbol{\pi}}(s,\boldsymbol{a}^0)^\top
\Delta\boldsymbol{a}_{\mathcal M}
+
o(\|\Delta\boldsymbol{a}_{\mathcal M}\|).
\end{equation}
\end{lemma}

where
$\Delta\boldsymbol{a}_{\mathcal M}
=\boldsymbol{a}_{\mathcal M}-\boldsymbol{a}_{\mathcal M}^{0}$
denotes the joint-action deviation of the compromised agents. Lemma~\ref{lem:cooperative_resistance} establishes the existence of Cooperative Resistance that locally offsets the safety cost increase induced
by compromised agents. The derivation is provided in
Appendix~C.

\section{MinMax Attack Framework}
\label{sec:method}
\begin{figure*}[ht]
  \centering 
  \small
  \includegraphics[width=0.9\linewidth]{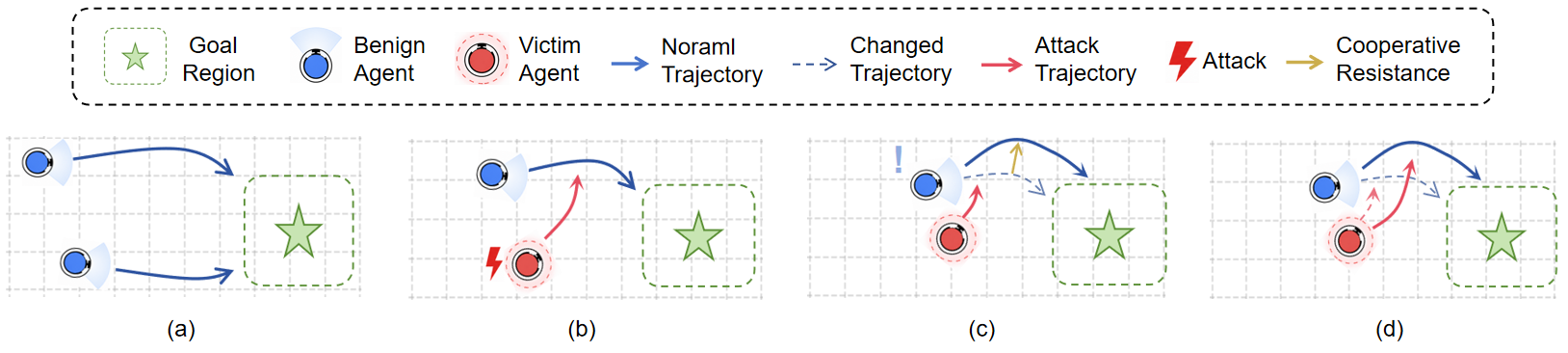} 
  \caption{MinMax Attack under Cooperative Resistance.
(a) Normal trajectories.
(b) The attack is initiated.
(c) Benign agents detect unsafe behaviors and perform Cooperative Resistance.
(d) The MinMax Attack accounts for Cooperative Resistance and adjusts.}
  \label{fig4}
\end{figure*}

Building on the multi-agent safety interactions established above, we propose MinMax Attack, an inference-time observation attack against SMARL.
MinMax Attack exploits NSSC through JOMAS, which jointly selects adversarial target actions for
compromised agents, while accounting for the Cooperative Resistance of benign agents during target actions optimization, as illustrated in Figure~\ref{fig4}. The selected target actions are then
realized through bounded observation perturbations, yielding a two-stage attack
framework.

\subsection{Two-Stage MinMax Attack}
 
As shown in Algorithm~\ref{alg1}, MinMax Attack consists of two stages.
In Stage I, the attack computes the nominal action
$a^i=\pi^i(s^i)$ for each agent and selects the compromised agents'
target actions through JOMAS. The Cooperative Resistance is denoted by
$\boldsymbol{\nu}_{\mathcal U}=\{\nu^j\}_{j\in\mathcal U}$, where each
$\nu^j$ is constrained by
$\|\nu^j\|_2\le\alpha\sigma^j$.
Here, $\alpha>0$ controls the response range and $\sigma^j$ denotes the
action standard deviation of benign agent $j$. The target-action objective is given in Eq.~\eqref{eq3}, where $Q^\pi$ denotes the critic used by the corresponding attack variant. Line~5 optimizes the bounded Cooperative Resistance $\boldsymbol{\nu}_{\mathcal U}$, while Line~7 performs JOMAS by jointly updating the compromised-agent target actions against the estimated Cooperative Resistance.

\begin{equation}
\label{eq3}
\begin{aligned}
\mathbf{a}_{\mathcal{M}}^\star
&=\arg\max_{\mathbf{a}_{\mathcal{M}}}\min_{\boldsymbol{\nu}_{\mathcal{U}}}
Q^\pi\!\left(s,\left(\mathbf{a}_{\mathcal{M}},\{a^j+\nu^j\}_{j\in\mathcal{U}}\right)\right)\\
&\quad\mathrm{s.t.}\quad \|\nu^j\|_2\le\alpha\sigma^j,\quad j\in\mathcal{U}.
\end{aligned}
\end{equation}

In Stage II, given the selected target action
$\boldsymbol{a}_{\mathcal M}^{\star}$, the attacker generates bounded
observation perturbations for the compromised agents. Each perturbation
$\delta^i$ is optimized such that the perturbed policy output
$\pi^i(s^i+\delta^i)$ approaches the target action $a^{i,\star}$.
After each update, $\delta^i$ is projected onto the
$\epsilon$-bounded feasible region, ensuring that the final adversarial
observation $\hat{s}^i=s^i+\delta^i$ satisfies the Perturbation Budget.

\begin{algorithm}[t]
\caption{Two-Stage MinMax Attack}
\label{alg1}
\noindent\textbf{Input:} State $s$, local policy inputs $\{s^i\}_{i\in\mathcal{N}}$, policies $\boldsymbol{\pi}$, critic $Q^\pi$.
\noindent\textbf{Parameters:} Attack iterations $K_a,K_\nu,K_\delta$, step sizes $\eta_a,\eta_\nu,\eta_\delta$.

\noindent\textbf{Output:} Adversarial observations $\{\hat{s}^i\}_{i\in\mathcal{M}}$.

\begin{algorithmic}[1]
\State Compute nominal actions $a^i \leftarrow \pi^i(s^i)$ for all $i\in\mathcal{N}$.
 
\State Initialize Cooperative Resistance deviations $\boldsymbol{\nu}_{\mathcal{U}} \leftarrow \mathbf{0}$ and compromised target actions
$\mathbf{a}_{\mathcal M}^{(0)}\leftarrow\{a^i\}_{i\in\mathcal{M}}$.

\Statex \textbf{Stage I: MinMax worst-action selection}
\For{$k=1,\ldots,K_a$}
    \For{$l=1,\ldots,K_\nu$}
        \State Update Cooperative Resistance by gradient descent:
        \[
        \boldsymbol{\nu}_{\mathcal{U}}
        \leftarrow
        \Pi_{\|\nu^j\|_2\le \alpha\sigma^j}
        \left[
        \boldsymbol{\nu}_{\mathcal{U}}
        -
        \eta_\nu
        \nabla_{\boldsymbol{\nu}_{\mathcal{U}}}
        Q^\pi\!\left(
        s,\mathbf{a}_{\mathcal{M}}^{(k-1)},\{a^j+\nu^j\}_{j\in\mathcal{U}}
        \right)
        \right].
        \]
    \EndFor
    \State Update compromised target actions by gradient ascent:
    \[
    \mathbf{a}_{\mathcal{M}}^{(k)}
    \leftarrow
    \Pi_{\mathcal{A}_{\mathcal{M}}}
    \left[
    \mathbf{a}_{\mathcal{M}}^{(k-1)}
    +
    \eta_a
    \nabla_{\mathbf{a}_{\mathcal{M}}}
    Q^\pi\!\left(
 s,\mathbf{a}_{\mathcal{M}}^{(k-1)},\{a_\mathcal{U}+\nu_\mathcal{U}\} 
    \right)
    \right].
    \]
\EndFor
\State Set $\mathbf{a}_{\mathcal{M}}^\star \leftarrow \mathbf{a}_{\mathcal{M}}^{(K_a)}$.

\Statex \textbf{Stage II: Observation perturbation generation}
\State Initialize $\delta^{i,(0)}\leftarrow 0$ for each compromised agent $i\in\mathcal{M}$.
\For{$k=1,\ldots,K_\delta$}
    \State Define the action-matching loss:
    \[
    \mathcal{L}_{\delta}
    =
    \sum_{i\in\mathcal{M}}
    \left\|
    \pi^i(s^i+\delta^{i,(k-1)})-a^{i,\star}
    \right\|_2^2 .
    \]
    \State Update perturbations by projected gradient descent:
    \[
    \delta^{i,(k)}
    \leftarrow
    \Pi_{\|\delta^i\|_p\le\epsilon}
    \left[
    \delta^{i,(k-1)}
    -
    \eta_\delta
    \nabla_{\delta^i}\mathcal{L}_{\delta}
    \right],
    \quad
    \forall i\in\mathcal{M}.
    \]
\EndFor
\State $\hat{s}^i=s^i+\delta^{i,(K_\delta)}$ for all $i\in\mathcal{M}$.
\State \Return $\{\hat{s}^i\}_{i\in\mathcal{M}}$.
\end{algorithmic}
\end{algorithm}

When $Q^\pi$ is instantiated as the reward critic, the MinMax-R variant promotes stealthiness by selecting adversarial actions that preserve high predicted task reward. The MinMax-C variant uses the cost critic and prioritizes safety cost maximization.

\subsection{SMARL Vulnerability Boundary Analysis}
\label{subsec:vulnerability-boundary}
We analyze the safety cost boundary under observation attacks and characterize its dependence on the Compromised-Agent Budget $m$ and the Perturbation Budget $\epsilon$. The resulting bounds quantify how much the safety cost can deviate from the clean policy under bounded perturbations, thereby providing a worst-case vulnerability envelope for SMARL.

We introduce the quantities used in the bound. Let $L_{\max}$ denote the maximum local Lipschitz constant of the agents' policies, and let $c_{\mathrm{ind}}$ denote an upper bound on the single-agent safety cost influence. Let $\mathcal{S}_c$ be the unsafe state set, and define
\begin{equation}
p_{\mathrm{unsafe}}(s)
=
\max_{\boldsymbol{a}}
\sum_{s'\in\mathcal{S}_c}
p(s'\mid s,\boldsymbol{a}),
\end{equation}
which is the maximum probability of transitioning from state $s$ to an unsafe state under any joint action. Let $V_c^{\boldsymbol{\pi}}(s)$ denote the clean safety value and $\tilde{V}_c^{\boldsymbol{\pi}}(s)$ denote the safety value under attack. We present the main upper bounds below and provide the complete derivations, together with the corresponding lower-bound analyses, in Appendix~A.

\textbf{Theorem 1} (One-Step Cost Bound).
The one-step safety cost deviation induced by attacking $m$ agents is bounded by
\begin{equation}
\label{eq:one-step-vulnerability-bound}
\tilde{V}_c^{\boldsymbol{\pi}}(s)
-
V_c^{\boldsymbol{\pi}}(s)
\le
2m^2\epsilon L_{\max}c_{\mathrm{ind}}
\left(
p_{\mathrm{unsafe}}(s)+\frac{\gamma}{1-\gamma}
\right).
\end{equation}
This result shows that the one-step safety cost upper bound scales quadratically with the Compromised-Agent Budget and linearly with the Perturbation Budget.  

\textbf{Theorem 2} (Episodic Cost Bound). For the episodic setting, assume that the clean policy satisfies the safety budget
$V_c^{\boldsymbol{\pi}}(\mu_0)\le \kappa$, where $\mu_0$ is the initial state distribution and $\kappa$ is the prescribed safety threshold. Under continuous observation attacks, the episodic safety cost satisfies
\begin{equation}
\label{eq:episodic-vulnerability-bound}
\begin{aligned}
V_c^{\tilde{\boldsymbol{\pi}}}(\mu_0)
&\le\kappa+2m^2\epsilon L_{\max}c_{\mathrm{ind}}\\
&\quad\times\left(\frac{1}{1-\gamma}+\frac{2\gamma m\epsilon L_{\max}}{(1-\gamma)^2}\right)\\
&\quad\times\left(\max_s p_{\mathrm{unsafe}}(s)+\frac{\gamma}{1-\gamma}\right).
\end{aligned}
\end{equation}
The episodic bound contains an $\mathcal{O}(m^2\epsilon)$ term from the one-step safety cost deviation and an additional $\mathcal{O}(m^3\epsilon^2)$ term induced by accumulated distribution shift.

\section{Experiments}
\subsection{Experimental Setup}
We conduct attack experiments on multiple SOTA SMARL algorithms across representative SMARL environments and compare the two MinMax Attack variants with representative attack baselines. All experiments instantiate the Perturbation
Budget with the $\ell_\infty$ norm. Following Eq.~\eqref{eq:attack_objective}, we regard an attack as stealthy when its mean episodic reward is no lower than the
clean reward, corresponding to $\eta_R=0$. Detailed experimental settings are provided in Appendices~D, E, and F.

\begin{figure}[ht]
  \centering
  \small
  \includegraphics[width=1.0\linewidth,
 height=2.1cm
  ]{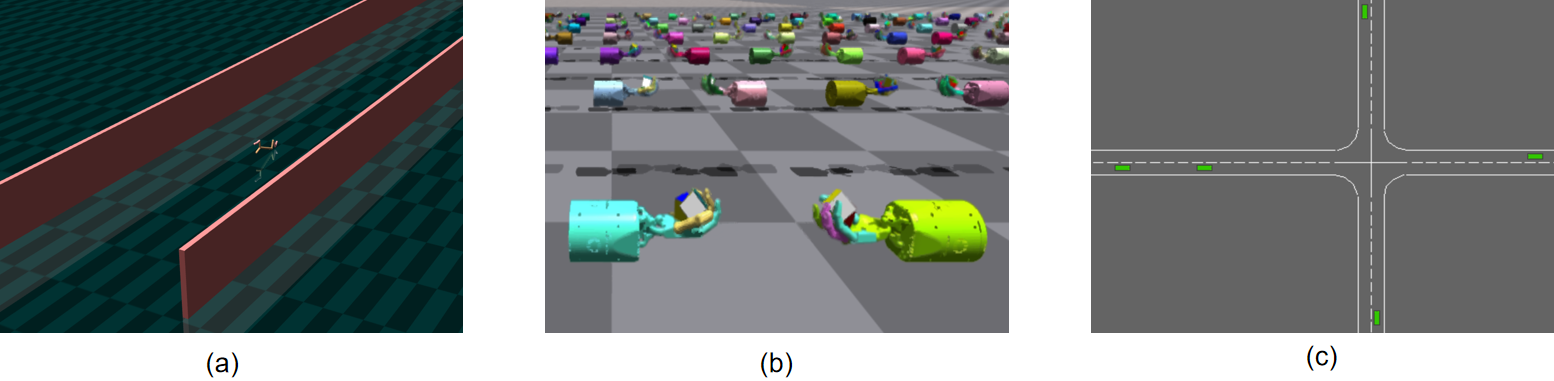} 
  \caption{(a) Safe Multi-Agent MuJoCo: HalfCheetah task. (b) Safe Multi-Agent Isaac Gym: ShadowHandReOrientation task. (c) Highway-env: Intersection task.}
  \label{fig1}
\end{figure}

\paragraph{Environments}
We evaluate the proposed inference-time observation attack on three safe multi-agent continuous-control benchmarks: Safe Multi-Agent MuJoCo~\cite{51_todorov2012mujoco, 26_gu2023safe}, Safe Multi-Agent Isaac Gym~\cite{50_makoviychuk2021isaac, 26_gu2023safe}, and Highway-env~\cite{52_leurent2018environment}, as illustrated in Figure~\ref{fig1}. These environments cover cooperative locomotion, dexterous manipulation, and autonomous driving, respectively, providing diverse interaction dynamics and safety constraints for evaluating attack effectiveness across different forms of multi-agent coordination.  

\paragraph{Victim SMARL Models}
We consider four representative SOTA SMARL victim models spanning both on-policy and off-policy learning paradigms: WCMASAC and HASAC-Lag~\cite{46_liu2026safe}, and MACPO and MAPPO-Lag~\cite{26_gu2023safe}. We evaluate both MinMax-R and MinMax-C under different Compromised-Agent Budgets and Perturbation Budgets, and report the cumulative reward and cumulative safety cost over each evaluation episode to jointly assess task performance based on reward and safety degradation based on safety cost.

\paragraph{Benchmark Observation Attacks}
We compare the proposed MinMax Attack against four representative observation attack baselines that cover different attack objectives. Random perturbation serves as an unstructured noise baseline without access to policy or critic information. Maximal Action Difference (MAD)~\cite{48_zhang2020robust} directly maximizes the deviation between nominal and perturbed actions, but does not explicitly optimize reward or safety cost. Max-Cost (MC) and Max-Reward (MR)~\cite{45_liu2022robustness} are critic-guided attacks that respectively maximize the predicted safety cost and task reward. Both have been shown to induce effective and reward-stealthy attacks in single-agent SRL, making them informative baselines for evaluating whether similar behavior persists in the multi-agent setting. These baselines allow us to compare MinMax Attack against generic perturbations, action-disruption attacks, and critic-guided safety- or reward-oriented attacks under the same observation attack setting.

\subsection{Attack Results Analysis}
Table~\ref{tab2} reports the attack performance on the six-agent ShadowHandReOrientation task across four victim Safe-MARL models and different numbers of attacked agents, with the Perturbation Budget set to $\epsilon=0.06$. Random and MAD neither consistently drive the victim into unsafe
regions nor preserve task reward sufficiently to remain stealthy.

\begin{table*}[ht]
\centering
\setlength{\tabcolsep}{1.0pt}         
\renewcommand{\arraystretch}{1.5}   %
\makebox[\textwidth][c]{%
\resizebox{\textwidth}{!}{%

\begin{tabular}{llcc|cc|cc|cc|cc|cc}
\toprule
Victim Model & Vic. &
\multicolumn{2}{c|}{\textbf{Random}} &
\multicolumn{2}{c|}{\textbf{MAD}} &
\multicolumn{2}{c|}{\textbf{MC}} &
\multicolumn{2}{c|}{\textbf{MR}} &
\multicolumn{2}{c|}{\textbf{MinMax-R (Ours)}} &
\multicolumn{2}{c}{\textbf{MinMax-C (Ours)}}\\
\cmidrule(lr){3-4}
\cmidrule(lr){5-6}
\cmidrule(lr){7-8}
\cmidrule(lr){9-10}
\cmidrule(lr){11-12}
\cmidrule(lr){13-14}
& & Reward $\uparrow$ & Cost $\downarrow$
& Reward $\uparrow$ & Cost $\downarrow$
& Reward $\uparrow$ & Cost $\downarrow$
& Reward $\uparrow$ & Cost $\downarrow$
& Reward $\uparrow$ & Cost $\downarrow$
& Reward $\uparrow$ & Cost $\downarrow$ \\
\midrule

\multirow{3}{*}{\shortstack{MACPO\\CR: 526.3\\ CC: 66.1}}
 
& 1 & 515.7 $\pm$ 54.9 & 95.8 $\pm$ 30.2
    & 500.6 $\pm$ 67.7 & 82.4$\pm$ 34.3
    & 524.0 $\pm$ 35.0& 96.9$\pm$ 29.4
    & 528.5 $\pm$ 62.3& 89.2 $\pm$ 22.7
    & 535.1 $\pm$ 68.9 & 95.4 $\pm$ 21.5
    & \cellcolor{gray!20}\textbf{530.2 $\pm$52.1}  & \cellcolor{gray!20}\textbf{106.5 $\pm$34.8}\\
& 2 & 504.6 $\pm$  51.9 & 107.2$\pm$ 43.0
    &511.0 $\pm$ 62.0 & 85.0 $\pm$ 27.5
    &536.0 $\pm$ 61.4& 119.3$\pm$ 39.3
    & 544.8 $\pm$ 61.2 & 100.4 $\pm$ 35.0
    & 552.0 $\pm$ 64.1 & 107.7 $\pm$  29.8
    & \cellcolor{gray!20}\textbf{544.8 $\pm$73.8} & \cellcolor{gray!20}\textbf{126.4 $\pm$ 40.8}\\
& 3 & 497.4 $\pm$ 55.1 & 128.8 $\pm$ 45.9
    &  507.3 $\pm$ 59.0 &  97.6 $\pm$ 42.0
    & 528.0 $\pm$ 51.5 & 122.6 $\pm$ 38.6
    & 550.7$\pm$ 70.1 & 117.9 $\pm$ 42.4
    & 567.2$\pm$ 57.6 & 122.5 $\pm$ 43.9 
    & \cellcolor{gray!20}\textbf{537.6 $\pm$67.2} & \cellcolor{gray!20}\textbf{129.8 $\pm$ 41.5} \\
\cmidrule(lr){1-14}

\multirow{3}{*}{\shortstack{MAPPO-Lag\\CR: 618.7\\ CC: 0.0}}
& 1 & 606.9 $\pm$ 38.8 & 0.0 $\pm$  0.0
    & 620.9 $\pm$ 45.7 & 0.0 $\pm$ 0.0
    & 622.7 $\pm$ 29.2 & 5.4 $\pm$  2.8
    & 635.7 $\pm$ 47.3 & 0.0 $\pm$ 0.0
    & 636.4 $\pm$ 37.4 & 0.5 $\pm$  0.2
    & \cellcolor{gray!20}\textbf{619.3$\pm$ 41.3} & \cellcolor{gray!20}\textbf{8.3 $\pm$ 4.3} \\
& 2 & 621.5 $\pm$ 33.3 & 0.0 $\pm$ 0.0
    & 633.6 $\pm$ 47.8 & 0.0$\pm$ 0.0
    & 621.8 $\pm$ 32.1 & 9.4$\pm$ 4.7
    & 643.2 $\pm$ 46.2 & 0.2 $\pm$0.1
    & 644.5 $\pm$ 45.3 & 1.9$\pm$ 1.0
    & \cellcolor{gray!20}\textbf{634.7$\pm$ 40.4} &  \cellcolor{gray!20}\textbf{10.7 $\pm$  5.6} \\
& 3 & 597.4 $\pm$ 46.5  & 0.0 $\pm$ 0.0
    & 614.7 $\pm$ 38.2 & 0.0 $\pm$ 0.0
    & 623.5$\pm$  41.0 & 18.6 $\pm$ 9.3
    & 641.0 $\pm$ 42.3 & 9.4 $\pm$ 2.5
    & 653.1 $\pm$ 34.7& 12.9 $\pm$2.8
    & \cellcolor{gray!20}\textbf{626.6 $\pm$ 37.3} & \cellcolor{gray!20}\textbf{21.4 $\pm$9.6} \\
\cmidrule(lr){1-14}

\multirow{3}{*}{\shortstack{WCMASAC\\CR: 632.9\\CC: 7.2}}
& 1 & 625.4 $\pm$ 45.4 & 8.8 $\pm$ 3.1
    & 624.7 $\pm$ 56.3 & 9.1$\pm$ 3.8
    & 634.1 $\pm$ 42.8 & 9.9 $\pm$ 2.9
    & 641.4 $\pm$ 43.2 & 8.8 $\pm$ 4.6
    & 647.5 $\pm$ 49.0& 10.2 $\pm$4.5
    & \cellcolor{gray!20}\textbf{640.0 $\pm$36.2}& \cellcolor{gray!20}\textbf{ 13.2 $\pm$4.4} \\
& 2 & 633.9$\pm$ 31.1 & 8.4 $\pm$ 3.2
    & 616.2$\pm$ 45.7& 8.9 $\pm$ 3.7
    &  630.5 $\pm$ 34.4& 16.1$\pm$  8.2
    &  651.8 $\pm$ 53.0 &  14.3 $\pm$6.1
    & \cellcolor{gray!20}\textbf{661.4 $\pm$ 51.9} & \cellcolor{gray!20}\textbf{21.5 $\pm$ 6.7}
    &635.7 $\pm$ 48.1 & 18.8 $\pm$ 8.6 \\
& 3 & 609.7 $\pm$ 46.0 & 8.1 $\pm$ 3.4
    & 626.0 $\pm$ 40.1 & 16.2 $\pm$ 7.2
    & 630.0 $\pm$ 36.5& 20.8$\pm$  8.8
    & 652.2 $\pm$ 42.5 & 15.8 $\pm$ 7.5
    & 660.5$\pm$ 49.3& 23.6 $\pm$ 7.2
    & \cellcolor{gray!20}\textbf{634.9 $\pm$ 37.5} & \cellcolor{gray!20}\textbf{26.3 $\pm$ 12.7}\\
\cmidrule(lr){1-14}
 
\multirow{3}{*}{\shortstack{HASAC-Lag\\CR: 542.7\\CC: 26.9}}
& 1 & 539.9 $\pm$29.7 & 37.4 $\pm$ 12.5
    & 529.1 $\pm$ 34.2 & 34.6 $\pm$ 8.4
    & 546.3 $\pm$ 33.6  & 33.8 $\pm$ 9.8  
    & 560.2 $\pm$ 42.1 & 38.0 $\pm$ 8.7 %
    &\cellcolor{gray!20}\textbf{562.1 $\pm$ 42.7} & \cellcolor{gray!20}\textbf{41.4 $\pm$ 7.5} %
    & 544.0 $\pm$ 30.8 & 35.3 $\pm$ 10.4 \\
& 2 & 525.6$\pm$ 34.9 & 34.5 $\pm$ 11.4
    & 523.3 $\pm$ 31.5 & 35.9 $\pm$ 6.6
    & 538.2 $\pm$ 26.8 & 41.1 $\pm$ 10.0
    & 577.4 $\pm$ 47.0 & 38.7 $\pm$ 9.2 %
    & 581.3 $\pm$ 43.4 & 40.5 $\pm$ 10.8 %
    & \cellcolor{gray!20}\textbf{548.0 $\pm$ 32.9} & \cellcolor{gray!20}\textbf{43.1 $\pm$ 11.8} \\
& 3 & 501.9 $\pm$ 48.3 & 46.9 $\pm$ 14.4
    & 531.6 $\pm$ 32.7 & 39.7 $\pm$ 7.6
    & 546.8 $\pm$ 29.9 & 42.2 $\pm$ 8.3
    & 574.3 $\pm$ 36.4 & 46.7 $\pm$ 10.3 %
    & \cellcolor{gray!20}\textbf{586.4$\pm$ 54.5}  & \cellcolor{gray!20}\textbf{50.1$\pm$ 12.2} %
    & 552.4$\pm$ 37.0 & 46.8 $\pm$ 11.5 \\
\midrule
\end{tabular}
}}
\caption{Attack performance on Safe Multi-Agent Isaac Gym: ShadowHandReOrientation (6$\times$2) under different victim models and varying numbers of attacked agents (Vic.). Bold values indicate the attack with the highest cost among methods whose reward is not lower than the baseline's clean reward (CR) within each row. CC denotes the clean cost.}
\label{tab2}
\end{table*} 
 
Critic-guided attacks induce larger cumulative costs. On MAPPO-Lag with three compromised agents, MinMax-C increases the cumulative cost to 21.4, approximately 15\% higher than the second-best MC attack, while maintaining a reward of 626.6, above the clean reward of 618.7. MinMax Attack also achieves the highest cost across the remaining victim algorithms and attack scales without causing reward collapse. Notably, reward and cost can increase simultaneously in SMARL, extending a phenomenon previously observed in single-agent safe RL~\cite{45_liu2022robustness} to the multi-agent setting. Under the same attack budget, MinMax-R induces a higher average cost than MR while preserving high reward, whereas MinMax-C better preserves reward than MC while producing stronger safety degradation. These results suggest that jointly optimizing compromised-agent actions and modeling Cooperative Resistance enable MinMax Attack to exploit multi-agent interactions more effectively than existing critic-guided attacks.

\begin{figure*}[ht]
  \small
  \centering
  \makebox[\linewidth][c]{%
  \includegraphics[width=0.8\linewidth,height=6.2cm]{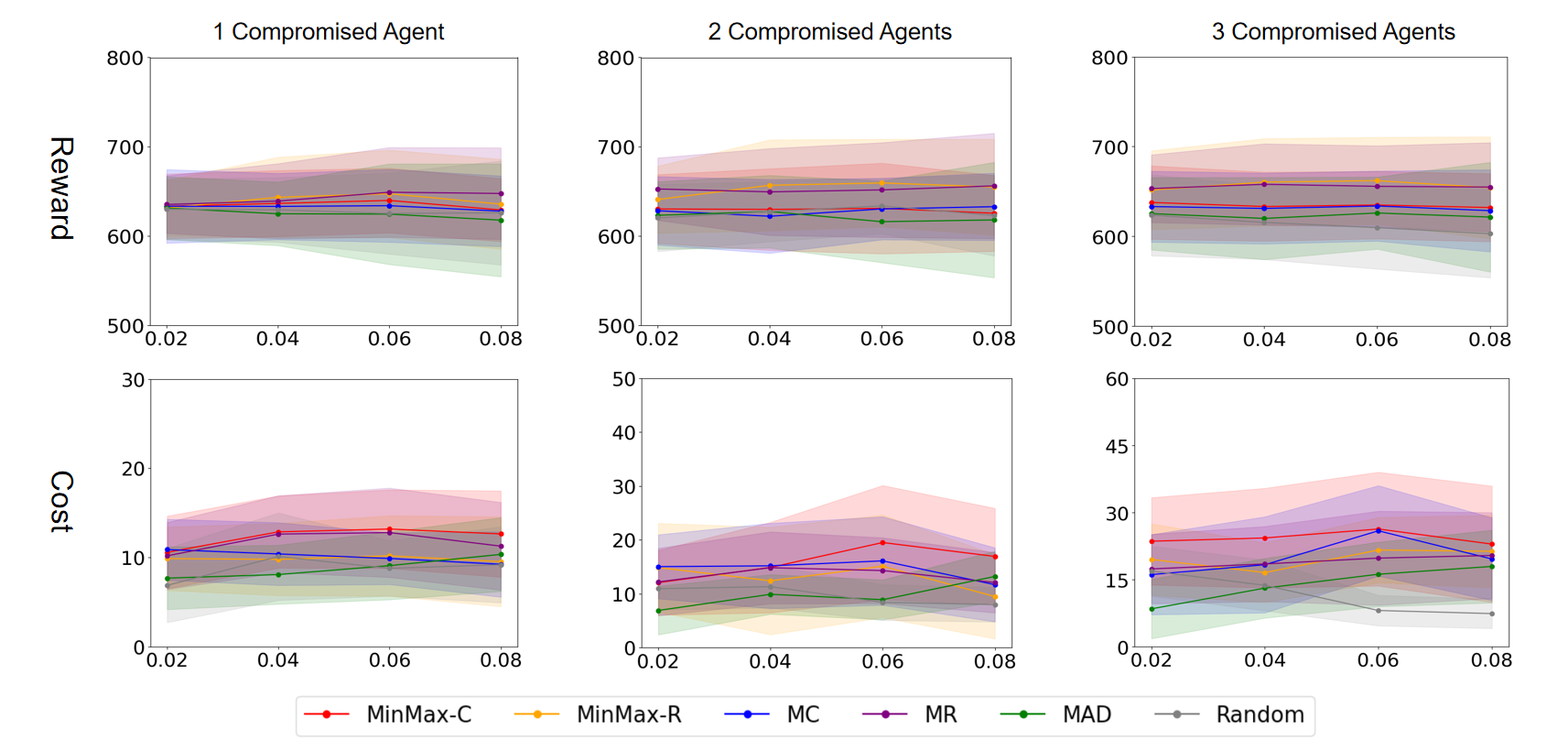}%
}
  \caption{Attack performance on WCMASAC under different numbers of compromised agents and Perturbation Budgets. The x-axis denotes the Perturbation Budget $\epsilon$.}
  \label{fig5}
\end{figure*}

As shown in Figure~\ref{fig5}, MinMax Attack consistently outperforms the baselines across different numbers of compromised agents. When two or three agents are compromised, MinMax-C and MinMax-R achieve their strongest attack performance around $\epsilon=0.06$. When only one agent is compromised, the most effective perturbation range is between $\epsilon=0.04$ and $\epsilon=0.06$. The results also show that increasing the Perturbation Budget does not always monotonically improve the attack.

We further report the computational efficiency of the attacks. Table~\ref{tab:app-runtime} presents the per-step runtime in Isaac Gym as mean $\pm$ standard deviation over ten runs. With two compromised agents, MC and MR require approximately $0.4$ seconds per attacked environment step, while MinMax-C and MinMax-R require approximately $0.6$ seconds. As the number of compromised agents increases to four, the runtime of MC and MR rises to about $0.7$ seconds, whereas both MinMax variants remain close to $0.6$ seconds. These results indicate that MinMax Attack incurs additional optimization overhead at small attack scales but exhibits more stable runtime as the Compromised-Agent Budget increases. 
\begin{table}[ht]
    \centering
    \caption{Per-step attack runtime in Isaac Gym. Results are reported in seconds per step.}
    \label{tab:app-runtime}
    \setlength{\tabcolsep}{4pt}
    \renewcommand{\arraystretch}{1.15}
    \resizebox{\columnwidth}{!}{%
\begin{tabular}{llccc}
        \toprule
        \multirow{2}{*}{Category} 
        & \multirow{2}{*}{Attack Method}
        & \multicolumn{3}{c}{Number of Compromised Agents $m$} \\
        \cmidrule(lr){3-5}
        & & $m=2$ & $m=3$ & $m=4$ \\
        \midrule
        \multirow{2}{*}{Existing Attacks}
        & MC       &  0.37 $\pm$ 0.04 & 0.56 $\pm$ 0.02 & 0.67 $\pm$ 0.02 \\
        & MR       & 0.40 $\pm$ 0.03 & 0.55 $\pm$ 0.02 & 0.68 $\pm$ 0.02 \\
        \midrule
        \multirow{2}{*}{MinMax Attack}
        & MinMax-C & 0.57 $\pm$ 0.03& 0.60 $\pm$ 0.03  & 0.59 $\pm$ 0.03 \\
        & MinMax-R  & 0.59 $\pm$ 0.02& 0.58 $\pm$ 0.02 & 0.58 $\pm$ 0.02 \\
        \bottomrule
    \end{tabular}%
}
\end{table}

\vspace{-5pt}
\subsection{Attack Scaling Analysis}
\begin{figure*}[h]
  \small
  \centering
  \makebox[\linewidth][c]{%
  \includegraphics[width=0.96\linewidth,height=6.4cm]{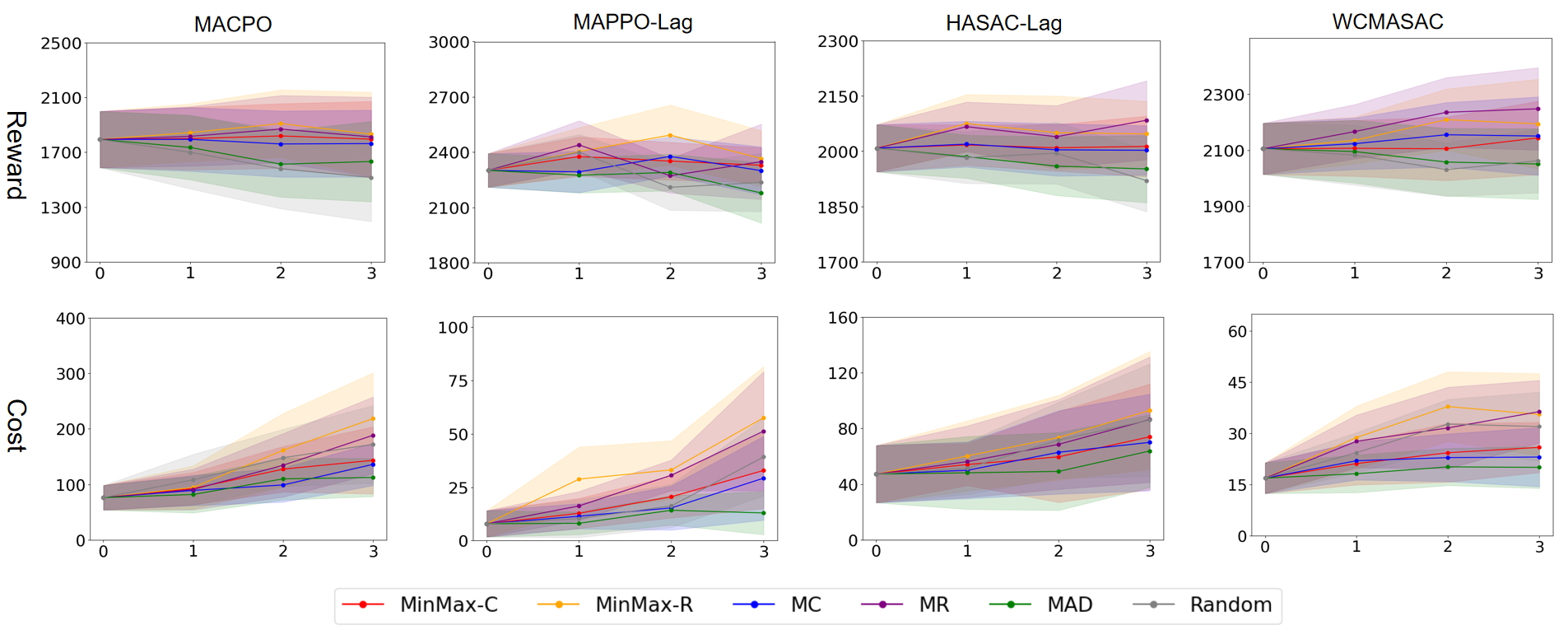}%
}
  \caption{Comparison of rewards and costs for different victim agents number across the four models. The x-axis denotes the Compromised-Agent Budget $m$.}
  \label{fig2}
\end{figure*}

Figure~\ref{fig2} evaluates how attack effectiveness scales with the Compromised-Agent Budget on the six-agent Safe Multi-Agent MuJoCo HalfCheetah task with the Perturbation Budget set to $\epsilon=0.08$. Across all four victim models, increasing the attacked subset consistently amplifies the cumulative safety cost. Among all methods, MinMax-R and MR exhibit the most pronounced cost escalation, with MinMax-R achieving the strongest attack effect across all four victim models. In contrast, MAD induces lower safety cost and shows less consistent scaling behavior as the Compromised-Agent Budget increases. In this environment, the reward-critic-guided attack induces a higher safety cost while maintaining task reward.

The impact on task reward is more mechanism-dependent. Compared with MAD and Random attacks, critic-guided attacks better preserve task reward while still
increasing safety cost, thereby improving stealthiness. By contrast, MAD and Random often cause a pronounced reward collapse when more agents are compromised, making their attacks easier to detect through task-performance degradation.  

\subsection{Ablation Study}
\begin{figure}[ht]
  \centering
  \includegraphics[width=0.82\linewidth,height=7.7cm]{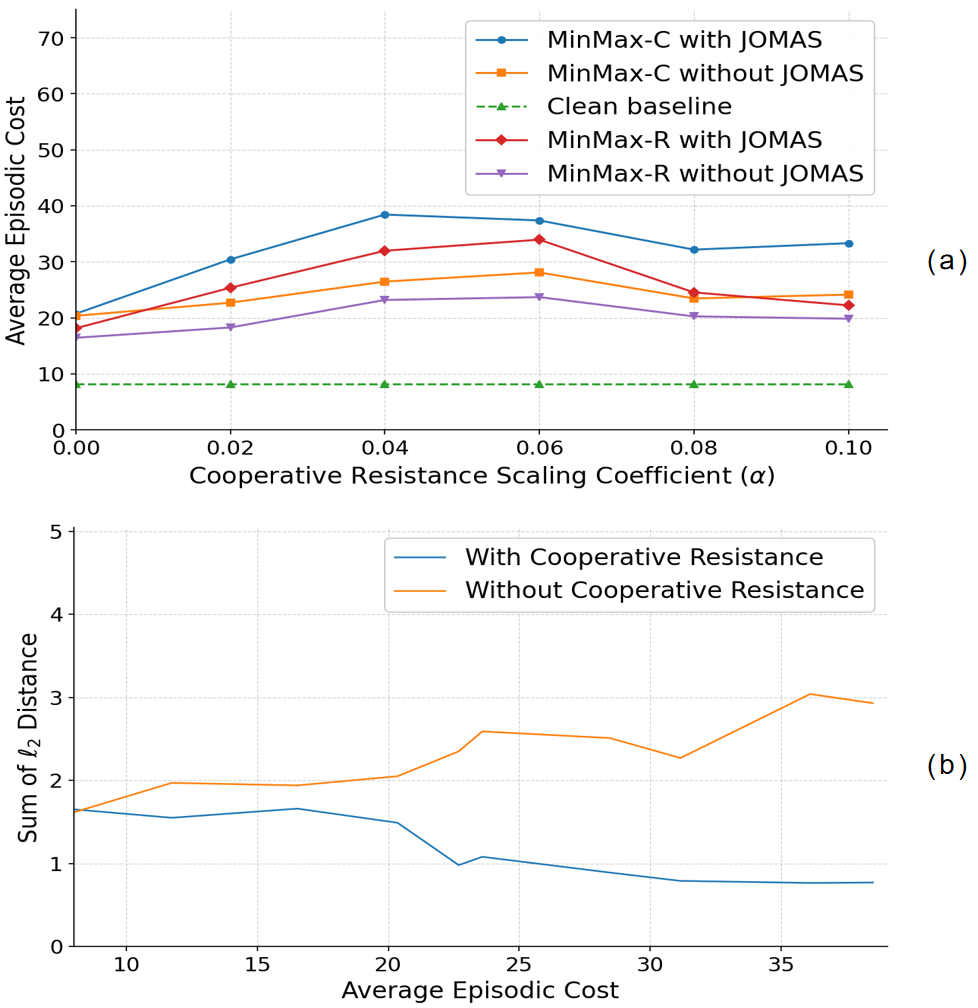}
  \caption{
(a) Impact of the Cooperative Resistance scaling coefficient $\alpha$ on the average episodic cost.
(b) Summed $\ell_2$ directional discrepancy between the action-change direction $\mathbf{a}_{t+1}-\mathbf{a}_{t}$ and the predicted Cooperative Resistance direction across benign agents under different cost levels.}
  \label{fig3}
\end{figure}

To isolate the effect of the JOMAS target-action selection mechanism, we conduct an ablation study on the six-agent Intersection task in Highway-Env. The victim policy is WCMASAC, $m=1, \epsilon=0.04$. This singleton-victim setting isolates the target-action selection mechanism from additional interactions among multiple compromised agents. We vary the Cooperative Resistance Scaling Coefficient $\alpha$ from $0$ to $0.1$ and evaluate both MinMax-C and MinMax-R under two settings: with JOMAS and without JOMAS. When JOMAS is disabled and $\alpha=0$, MinMax-C and MinMax-R reduce to the corresponding MC and MR attack variants, respectively. As shown in Figure~\ref{fig3}(a), incorporating JOMAS consistently increases the average episodic cost compared with the non-JOMAS variants, indicating that the target-action selection mechanism strengthens safety cost increase even under a single compromised agent. The improvement is particularly pronounced for MinMax-C, which achieves higher costs than its non-JOMAS counterpart across most values of $\alpha$. The attack effect is strongest when $\alpha$ lies in the range of $0.04$--$0.06$, suggesting that moderate cooperative-resistance modeling provides the most effective adversarial target selection.

We further evaluate whether the Cooperative Resistance module in MinMax-C can anticipate the compensatory behavior of benign agents. As shown in Figure~\ref{fig3}(b), incorporating cooperative resistance reduces the summed normalized $\ell_2$ directional discrepancy compared with the variant without Cooperative Resistance. As the average episodic cost increases, the action variation between consecutive time steps becomes larger, suggesting stronger Cooperative Resistance from benign agents. In this high-cost regime, Cooperative Resistance provides a more pronounced correction effect, leading to a smaller discrepancy between the predicted resistance behavior and the actual next-step action changes.  

We further conduct a parameter-sensitivity ablation to examine whether the performance of MinMax Attack depends critically on its optimization settings. The default attack iteration counts and step sizes follow the optimization settings of the MC and MR baselines~\cite{45_liu2022robustness}. Table~\ref{tab:app-iterations} varies the iteration counts with two compromised agents, where $K_a$ controls adversarial target-action updates, $K_{\nu}$ controls Cooperative Resistance updates, and $K_{\delta}$ controls observation-perturbation updates, as defined in Algorithm~\ref{alg1}. The results show only moderate sensitivity to these parameters, and increasing the iteration counts does not consistently yield substantial performance gains. 

\begin{table}[t]
\centering
\caption{Sensitivity of MinMax-C and MinMax-R to optimization iterations
with $m=2$.}
\label{tab:app-iterations}
\small
\setlength{\tabcolsep}{5pt}
\resizebox{\columnwidth}{!}{%
\begin{tabular}{llccccc}
\toprule
\multirow{2}{*}{Victim}
& \multirow{2}{*}{Attack}
& \multicolumn{3}{c}{Iterations}
& \multirow{2}{*}{Reward}
& \multirow{2}{*}{Cost} \\
\cmidrule(lr){3-5}
& & $K_a$ & $K_{\nu}$ & $K_{\delta}$ & & \\
\midrule

\multirow{6}{*}{WCMASAC}
& \multirow{3}{*}{MinMax-C}
& 1  & 5  & 5  & $628.7 \pm 49.1$ & $18.3 \pm 9.6$ \\
& & 5  & 10 & 10 & $631.1 \pm 50.7$ & $19.5 \pm 10.6$ \\
& & 10 & 15 & 15 & $632.8 \pm 51.6$ & $20.9 \pm 11.2$ \\
\cmidrule(lr){2-7}
& \multirow{3}{*}{MinMax-R}
& 1  & 5  & 5  & $651.3 \pm 48.9$ & $14.5 \pm 8.4$ \\
& & 5  & 10 & 10 & $659.6 \pm 52.6$ & $15.1 \pm 9.5$ \\
& & 10 & 15 & 15 & $659.3 \pm 49.9$ & $15.4 \pm 8.8$ \\
\midrule

\multirow{6}{*}{MACPO}
& \multirow{3}{*}{MinMax-C}
& 1  & 5  & 5  & $548.6 \pm 70.4$ & $119.4 \pm 40.9$ \\
& & 5  & 10 & 10 & $544.8 \pm 73.8$ & $126.4 \pm 40.8$ \\
& & 10 & 15 & 15 & $542.8 \pm 67.2$ & $125.7 \pm 41.4$ \\
\cmidrule(lr){2-7}
& \multirow{3}{*}{MinMax-R}
& 1  & 5  & 5  & $547.1 \pm 60.9$ & $103.9 \pm 26.2$ \\
& & 5  & 10 & 10 & $552.0 \pm 64.1$ & $107.7 \pm 29.8$ \\
& & 10 & 15 & 15 & $554.5 \pm 60.0$ & $109.4 \pm 28.0$ \\
\bottomrule
\end{tabular}%
}
\end{table}
\vspace{-5pt}
\section{Discussion}

\paragraph{Limitations.}
First, MinMax Attack assumes white-box access to the deployed policy and critic models. A practical extension to gray-box settings is to learn surrogate policies or critics from observed trajectories and exploit transferability, while black-box variants could use query-based or zeroth-order optimization to approximate the required gradients~\cite{68_pan2019characterizing,69_zhao2020blackbox,70_yu2022natural}. These extensions would reduce the dependence on internal model access, but may introduce surrogate approximation error or additional query overhead. Second, the current implementation relies on differentiable actor--critic architectures to optimize both adversarial target actions and observation perturbations. For non-differentiable controllers or systems without accessible critics, the gradient-based optimization could be replaced by derivative-free search over candidate actions and perturbations, or by learning differentiable surrogate policy and value models from interaction trajectories~\cite{71_chen2017zoo,69_zhao2020blackbox}. Evaluating these extensions in SMARL remains an important direction for future work.

\paragraph{Potential Defenses.}
Our results suggest that defending SMARL requires monitoring safety-related behavior in addition to task performance \cite{49_sun2021strongest,59_zhang2021robust}. Prior work has shown that adversarial regimes in RL can be detected online through inconsistencies in policy-value signals, such as advantage estimates~\cite{64_havens2018online}. This motivates safety-aware consistency monitoring in SMARL, where predicted safety cost can be compared with observed cost, or jointly monitored with task reward to identify anomalous cases in which reward remains nominal while safety cost increases. A complementary direction is robust policy learning. State-adversarial policy regularization has been shown to improve robustness against bounded observation perturbations in single-agent RL~\cite{48_zhang2020robust}. Extending such training to coordinated multi-agent perturbations may therefore reduce the vulnerability exposed by MinMax Attack.

\section{Conclusion}
This paper investigates the vulnerability of SMARL to inference-time observation attacks. We propose MinMax Attack, it jointly selects adversarial target actions for compromised agents while accounting for the Cooperative Resistance of benign agents. Our results show that cooperative mechanisms designed to maintain safety can become exploitable attack surfaces, enabling stronger safety degradation than existing observation attack baselines in the evaluated settings. We further provide a theoretical analysis characterizing how the induced safety cost upper bound scales with the Compromised-Agent Budget and the Perturbation Budget. Extensive experiments across representative SMARL environments and algorithms demonstrate that MinMax Attack is effective and stealthy, inducing substantial safety violations while preserving nominal task performance.

\bibliographystyle{ACM-Reference-Format}
\bibliography{ref}

\clearpage
\appendix
\section*{Appendix}

\phantomsection
\addcontentsline{toc}{section}{Appendix}

\begingroup
\small
\etocsetnexttocdepth{subsection}
\etocsetlocaltop{part}
\etocsettocstyle{}{}
\localtableofcontents
\endgroup

\section{Proof of  Theorems}
\label{appA}

We consider a Multi-Agent Constrained Markov Game with $n$ agents. Let
$s\in\mathcal{S}$ denote the global state and $\mathcal{A}$
denote the joint action space. The joint action is denoted by
$\boldsymbol{a}=(a^1,\ldots,a^n)$. For each agent $i$, let $s^i$ denote its local policy input associated with the global state $s$. During inference, all policies are fixed, and
the decentralized joint policy factorizes as
\begin{equation}
\boldsymbol{\pi}(\mathbf{a}\mid s)
=
\prod_{i=1}^{n}\pi^i(a^i\mid s^i).
\end{equation}
An adversary compromises a subset of agents
$\mathcal{M}\subseteq\mathcal{N}=\{1,\ldots,n\}$ with cardinality
$|\mathcal{M}|=m\le n$.
The complementary subset of unattacked agents is
$\mathcal{U}=\mathcal{N}\setminus\mathcal{M}$. For each attacked agent
$i\in\mathcal{M}$, the adversary perturbs the local policy input from $s^i$ to
$\tilde{s}^i$, where $\tilde{s}^i\in B_p^\epsilon(s^i)$ (an $\ell_p$-ball of radius $\epsilon$). For each unattacked
agent $j\in\mathcal{U}$, the observation remains unperturbed. Hence, the attacked joint policy is
\begin{equation}
\begin{aligned}
\tilde{\boldsymbol{\pi}}(\boldsymbol{a}\mid s)
&=\tilde{\boldsymbol{\pi}}^{\mathcal{M}}(\boldsymbol{a}^{\mathcal{M}}\mid s)
\boldsymbol{\pi}^{\mathcal{U}}(\boldsymbol{a}^{\mathcal{U}}\mid s)\\
&=\left(\prod_{i\in\mathcal{M}}\pi^i(a^i\mid\tilde{s}^i)\right)
\left(\prod_{j\in\mathcal{U}}\pi^j(a^j\mid s^j)\right).
\end{aligned}
\end{equation}

The $\pi$ is the clean policy. $\gamma\in[0,1)$ is the discount factor, which determines how future safety cost is weighted relative to the immediate cost. We define the cost value function and the cost Q-function. 
For a current state $s$ and action $a$, the next state is denoted by $s'$, which is sampled from the environment transition kernel
$p(\cdot\mid s,a)$. Throughout this section $c(s,a,s')$ denotes the instantaneous safety cost incurred when the system transitions from $s$ to $s'$ after executing action $a$. The cost value function $V_c^\pi(s)$ denotes the expected discounted cumulative safety cost starting from state $s$ and then following the clean policy $\pi$. Formally,
\begin{equation}
    V_c^\pi(s)
    =
    \mathbb{E}_{a_t\sim\pi(\cdot\mid s_t),
    s_{t+1}\sim p(\cdot\mid s_t,a_t)}
    \left[
    \sum_{t=0}^{\infty}\gamma^t
    c(s_t,a_t,s_{t+1})
    \,\middle|\, s_0=s
    \right].
\end{equation}

Accordingly, the clean cost Q-function under the fixed joint policy
$\boldsymbol{\pi}$ is defined as
\begin{equation}
    Q_c^{\boldsymbol{\pi}}(s,\boldsymbol{a})
    =
    \mathbb{E}_{s'\sim p(\cdot\mid s,\boldsymbol{a})}
    \left[
    c(s,\boldsymbol{a},s')
    +
    \gamma V_c^{\boldsymbol{\pi}}(s')
    \right],
\end{equation}
where $s'$ is the next state sampled from the transition kernel
$p(\cdot\mid s,\boldsymbol{a})$, $c(s,\boldsymbol{a},s')$ is the instantaneous
safety cost, $V_c^{\boldsymbol{\pi}}(s')$ is the expected discounted future
safety cost under the clean joint policy $\boldsymbol{\pi}$, and
$\gamma\in[0,1)$ is the discount factor.

We do not assume that the safety cost critic is additively separable across
agents. Instead, we assume bounded single-agent action influence. For any
agent $i$, state $s$, fixed actions $\mathbf a^{-i}$ of the other agents,
and two actions $a^i,\bar a^i\in\mathcal A_i$,
\begin{equation}
\left|
Q_c^{\boldsymbol{\pi}}(s,a^i,\mathbf a^{-i})
-
Q_c^{\boldsymbol{\pi}}(s,\bar a^i,\mathbf a^{-i})
\right|
\le B_c(s),
\end{equation}
where $B_c(s)$ bounds the largest critic variation caused by changing only
one agent's action.

\subsection{Proof of Theorem1}
\label{appA1}
The fixed inference policy of each agent is locally Lipschitz continuous with
respect to total variation distance. Specifically, for any attacked agent
$i\in\mathcal{M}$,
\begin{equation}
D_{\mathrm{TV}}
\left[
\pi^i(\cdot\mid \tilde{s}^i)
\,\middle\|\,
\pi^i(\cdot\mid s^i)
\right]
\le
L_i\|\tilde{s}^i-s^i\|_p
\end{equation}
where $L_{max}=\max_{1\le i\le n}L_i$.

Since the joint policy is a product measure and the agents in $\mathcal{U}$
are unperturbed, the joint total variation distance is bounded by the sum of
the individual policy shifts over the attacked agents:
\begin{align}
    &D_{\mathrm{TV}}
    \left[
    \tilde{\boldsymbol{\pi}}(\cdot\mid s)
    \,\middle\|\,
    \boldsymbol{\pi}(\cdot\mid s)
    \right] \nonumber\\
    &=
    D_{\mathrm{TV}}
    \left[
    \tilde{\boldsymbol{\pi}}^{\mathcal{M}}(\cdot\mid s)
    \,\middle\|\,
    \boldsymbol{\pi}^{\mathcal{M}}(\cdot\mid s)
    \right] \nonumber\\
    &\le
    \sum_{i\in\mathcal{M}}
    D_{\mathrm{TV}}
    \left[
    \pi^i(\cdot\mid \tilde{s}^i)
    \,\middle\|\,
    \pi^i(\cdot\mid s^i)
    \right]
    \le
    mL_{max}\epsilon .
\end{align}

Let $\mathcal{S}_c$ denote the unsafe state set. $p_s$ denote the maximum probability of transitioning into an unsafe state under the worst-case action.
\begin{equation}
    p_s
    =
    \max_{\boldsymbol{a}\in\boldsymbol{\mathcal{A}}}
    \sum_{s'\in\mathcal{S}_c}p(s'\mid s,\boldsymbol{a}).
\end{equation}

We assume that the single-agent critic influence satisfies
\begin{equation}
B_c(s)
\le
c_{ind}
\left(
p_s+\frac{\gamma}{1-\gamma}
\right),
\end{equation}
where $c_{ind}$ is the corresponding single-agent influence bound.

We define the clean-context cost Q-function of the compromised agents as
\begin{equation}
    Q_c^{\boldsymbol{\pi},\mathcal{M}}
    (s,\boldsymbol{a}^{\mathcal{M}})
    =
    \mathbb{E}_{\boldsymbol{a}^{\mathcal{U}}\sim
    \boldsymbol{\pi}^{\mathcal{U}}(\cdot\mid s)}
    \left[
    Q_c^{\boldsymbol{\pi}}
    (s,\boldsymbol{a}^{\mathcal{M}},\boldsymbol{a}^{\mathcal{U}})
    \right].
\end{equation}

For any two compromised-agent joint actions
$\boldsymbol{a}^{\mathcal{M}}$ and
$\bar{\boldsymbol{a}}^{\mathcal{M}}$, changing the actions of the
$m$ compromised agents one at a time and applying the bounded
single-agent influence condition gives
\begin{equation}
    \left|
    Q_c^{\boldsymbol{\pi},\mathcal{M}}
    (s,\boldsymbol{a}^{\mathcal{M}})
    -
    Q_c^{\boldsymbol{\pi},\mathcal{M}}
    (s,\bar{\boldsymbol{a}}^{\mathcal{M}})
    \right|
    \le
    m B_c(s).
\end{equation}

Define the one-step attacked cost surrogate as
\begin{equation}
    \tilde{V}_c^{\boldsymbol{\pi},\boldsymbol{\nu}}(s)
    =
    \mathbb{E}_{\boldsymbol{a}\sim\tilde{\boldsymbol{\pi}}(\cdot\mid s)}
    \left[
    Q_c^{\boldsymbol{\pi}}(s,\boldsymbol{a})
    \right].
\end{equation}
By the Bellman equation,
\begin{equation}
    V_c^{\boldsymbol{\pi}}(s)
    =
    \mathbb{E}_{\boldsymbol{a}\sim\boldsymbol{\pi}(\cdot\mid s)}
    \left[
    Q_c^{\boldsymbol{\pi}}(s,\boldsymbol{a})
    \right].
\end{equation}
Therefore,
\begin{align}
    &\tilde{V}_c^{\boldsymbol{\pi},\boldsymbol{\nu}}(s)
    -
    V_c^{\boldsymbol{\pi}}(s) \nonumber\\
    &=
    \sum_{\boldsymbol{a}\in\boldsymbol{\mathcal{A}}}
    \left(
    \tilde{\boldsymbol{\pi}}(\boldsymbol{a}\mid s)
    -
    \boldsymbol{\pi}(\boldsymbol{a}\mid s)
    \right)
    Q_c^{\boldsymbol{\pi}}(s,\boldsymbol{a}) \nonumber\\
    &=
    \sum_{\boldsymbol{a}^{\mathcal{M}}}
    \sum_{\boldsymbol{a}^{\mathcal{U}}}
    \left(
    \tilde{\boldsymbol{\pi}}^{\mathcal{M}}
    (\boldsymbol{a}^{\mathcal{M}}\mid s)
    -
    \boldsymbol{\pi}^{\mathcal{M}}
    (\boldsymbol{a}^{\mathcal{M}}\mid s)
    \right)
    \boldsymbol{\pi}^{\mathcal{U}}
    (\boldsymbol{a}^{\mathcal{U}}\mid s)
    Q_c^{\boldsymbol{\pi}}
    (s,\boldsymbol{a}^{\mathcal{M}},\boldsymbol{a}^{\mathcal{U}}) \nonumber\\
    &=
    \sum_{\boldsymbol{a}^{\mathcal{M}}}
    \left(
    \tilde{\boldsymbol{\pi}}^{\mathcal{M}}
    (\boldsymbol{a}^{\mathcal{M}}\mid s)
    -
    \boldsymbol{\pi}^{\mathcal{M}}
    (\boldsymbol{a}^{\mathcal{M}}\mid s)
    \right)
    Q_c^{\boldsymbol{\pi},\mathcal{M}}
    (s,\boldsymbol{a}^{\mathcal{M}}).
\end{align}
Let $\boldsymbol{a}^{\mathcal{M},0}$ be an arbitrary fixed reference
action of the compromised agents. Since both
$\tilde{\boldsymbol{\pi}}^{\mathcal{M}}(\cdot\mid s)$ and
$\boldsymbol{\pi}^{\mathcal{M}}(\cdot\mid s)$ are probability
distributions,
\begin{align}
    &\tilde{V}_c^{\boldsymbol{\pi},\boldsymbol{\nu}}(s)
    -
    V_c^{\boldsymbol{\pi}}(s) \nonumber\\
    &=
    \sum_{\boldsymbol{a}^{\mathcal{M}}}
    \left(
    \tilde{\boldsymbol{\pi}}^{\mathcal{M}}
    (\boldsymbol{a}^{\mathcal{M}}\mid s)
    -
    \boldsymbol{\pi}^{\mathcal{M}}
    (\boldsymbol{a}^{\mathcal{M}}\mid s)
    \right)
    \left[
    Q_c^{\boldsymbol{\pi},\mathcal{M}}
    (s,\boldsymbol{a}^{\mathcal{M}})
    -
    Q_c^{\boldsymbol{\pi},\mathcal{M}}
    (s,\boldsymbol{a}^{\mathcal{M},0})
    \right].
\end{align}

Therefore,
\begin{align}
    &\tilde{V}_c^{\boldsymbol{\pi},\boldsymbol{\nu}}(s)
    -
    V_c^{\boldsymbol{\pi}}(s) \nonumber\\
    &\le
    \sum_{\boldsymbol{a}^{\mathcal{M}}}
    \left|
    \tilde{\boldsymbol{\pi}}^{\mathcal{M}}
    (\boldsymbol{a}^{\mathcal{M}}\mid s)
    -
    \boldsymbol{\pi}^{\mathcal{M}}
    (\boldsymbol{a}^{\mathcal{M}}\mid s)
    \right|
    \max_{\boldsymbol{a}^{\mathcal{M}}}
    \left|
    Q_c^{\boldsymbol{\pi},\mathcal{M}}
    (s,\boldsymbol{a}^{\mathcal{M}})
    -
    Q_c^{\boldsymbol{\pi},\mathcal{M}}
    (s,\boldsymbol{a}^{\mathcal{M},0})
    \right|.
\end{align}

Using the definition of total variation distance,
\begin{align}
    &\tilde{V}_c^{\boldsymbol{\pi},\boldsymbol{\nu}}(s)
    -
    V_c^{\boldsymbol{\pi}}(s) \nonumber\\
    &\le
    2D_{\mathrm{TV}}
    \left[
    \tilde{\boldsymbol{\pi}}^{\mathcal{M}}(\cdot\mid s)
    \,\middle\|\,
    \boldsymbol{\pi}^{\mathcal{M}}(\cdot\mid s)
    \right]
    \max_{\boldsymbol{a}^{\mathcal{M}}}
    \left|
    Q_c^{\boldsymbol{\pi},\mathcal{M}}
    (s,\boldsymbol{a}^{\mathcal{M}})
    -
    Q_c^{\boldsymbol{\pi},\mathcal{M}}
    (s,\boldsymbol{a}^{\mathcal{M},0})
    \right|.
\end{align}

Using the joint TV bound,
\begin{equation}
    D_{\mathrm{TV}}
    \left[
    \tilde{\boldsymbol{\pi}}^{\mathcal{M}}(\cdot\mid s)
    \,\middle\|\,
    \boldsymbol{\pi}^{\mathcal{M}}(\cdot\mid s)
    \right]
    \le
    mL_{max}\epsilon.
\end{equation}

Moreover, by the bounded single-agent influence condition,
\begin{align}
    &\max_{\boldsymbol{a}^{\mathcal{M}}}
    \left|
    Q_c^{\boldsymbol{\pi},\mathcal{M}}
    (s,\boldsymbol{a}^{\mathcal{M}})
    -
    Q_c^{\boldsymbol{\pi},\mathcal{M}}
    (s,\boldsymbol{a}^{\mathcal{M},0})
    \right|
    \nonumber\\
    &\le
    m B_c(s)
    \le
    m c_{ind}
    \left(
    p_s+\frac{\gamma}{1-\gamma}
    \right).
\end{align}

Combining the two inequalities yields
\begin{align}
    \tilde{V}_c^{\boldsymbol{\pi},\boldsymbol{\nu}}(s)
    -
    V_c^{\boldsymbol{\pi}}(s)
    &\le
    2
    \left(
    mL_{max}\epsilon
    \right)
    \left[
    m c_{ind}
    \left(
    p_s+\frac{\gamma}{1-\gamma}
    \right)
    \right] \nonumber\\
    &=
    2m^2\epsilon L_{max}c_{ind}
    \left(
    p_s+\frac{\gamma}{1-\gamma}
    \right).
\end{align}

Using the same decomposition and bounds derived above, we can similarly lower-bound the one-step safety cost deviation as
\begin{align}
\tilde{V}_c^{\boldsymbol{\pi},\boldsymbol{\nu}}(s)
-
V_c^{\boldsymbol{\pi}}(s)
&\ge
-
\sum_{\boldsymbol{a}^{\mathcal{M}}}
\left|
\tilde{\boldsymbol{\pi}}^{\mathcal{M}}
(\boldsymbol{a}^{\mathcal{M}}\mid s)
-
\boldsymbol{\pi}^{\mathcal{M}}
(\boldsymbol{a}^{\mathcal{M}}\mid s)
\right|
\max_{\boldsymbol{a}^{\mathcal{M}}}
\sum_{i\in\mathcal{M}}
Q_{c,i}^{\boldsymbol{\pi}}(s,a^i) \nonumber\\
&=
-2D_{\mathrm{TV}}
\left[
\tilde{\boldsymbol{\pi}}^{\mathcal{M}}(\cdot\mid s)
\,\middle\|\,
\boldsymbol{\pi}^{\mathcal{M}}(\cdot\mid s)
\right]
\max_{\boldsymbol{a}^{\mathcal{M}}}
\sum_{i\in\mathcal{M}}
Q_{c,i}^{\boldsymbol{\pi}}(s,a^i).
\end{align}
Using the same bounds
$D_{\mathrm{TV}}\le mL_{max}\epsilon$ and
$\max_{\boldsymbol{a}^{\mathcal{M}}}\sum_{i\in\mathcal{M}}
Q_{c,i}^{\boldsymbol{\pi}}(s,a^i)
\le mc_{ind}\left(p_s+\frac{\gamma}{1-\gamma}\right)$,
we obtain
\begin{equation}
\tilde{V}_c^{\boldsymbol{\pi},\boldsymbol{\nu}}(s)
-
V_c^{\boldsymbol{\pi}}(s)
\ge
-2m^2\epsilon L_{max}c_{ind}
\left(
p_s+\frac{\gamma}{1-\gamma}
\right).
\end{equation}

\subsection{Proof of Theorem2}
\label{appA2}

We now extend the one-step cost bound to the discounted episodic cost. Let the
clean cost advantage be
\begin{equation}
    A_c^{\boldsymbol{\pi}}(s,\boldsymbol{a})
    =
    Q_c^{\boldsymbol{\pi}}(s,\boldsymbol{a})
    -
    V_c^{\boldsymbol{\pi}}(s).
\end{equation}
Define
\begin{equation}
    \delta_c^{\tilde{\boldsymbol{\pi}}}
    =
    \max_s
    \left|
    \mathbb{E}_{\boldsymbol{a}\sim
    \tilde{\boldsymbol{\pi}}(\cdot\mid s)}
    A_c^{\boldsymbol{\pi}}(s,\boldsymbol{a})
    \right|.
\end{equation}
By the performance difference bound for joint policies,
\begin{equation}
\label{eq:joint-cpo-bound}
    V_c^{\tilde{\boldsymbol{\pi}}}(\mu_0)
    -
    V_c^{\boldsymbol{\pi}}(\mu_0)
    \le
    \frac{\delta_c^{\tilde{\boldsymbol{\pi}}}}{1-\gamma}
    +
    \frac{2\gamma\delta_c^{\tilde{\boldsymbol{\pi}}}}{(1-\gamma)^2}
    \max_s
    D_{\mathrm{TV}}
    \left[
    \tilde{\boldsymbol{\pi}}(\cdot\mid s)
    \,\middle\|\,
    \boldsymbol{\pi}(\cdot\mid s)
    \right].
\end{equation}

For any state $s$, the expected clean advantage under the attacked joint policy
is
\begin{align}
    &\mathbb{E}_{\boldsymbol{a}\sim
    \tilde{\boldsymbol{\pi}}(\cdot\mid s)}
    A_c^{\boldsymbol{\pi}}(s,\boldsymbol{a}) \nonumber\\
    &=
    \mathbb{E}_{\boldsymbol{a}\sim
    \tilde{\boldsymbol{\pi}}(\cdot\mid s)}
    \left[
    Q_c^{\boldsymbol{\pi}}(s,\boldsymbol{a})
    -
    V_c^{\boldsymbol{\pi}}(s)
    \right] \nonumber\\
    &=
    \mathbb{E}_{\boldsymbol{a}\sim
    \tilde{\boldsymbol{\pi}}(\cdot\mid s)}
    Q_c^{\boldsymbol{\pi}}(s,\boldsymbol{a})
    -
    V_c^{\boldsymbol{\pi}}(s) \nonumber\\
    &=
    \tilde{V}_c^{\boldsymbol{\pi},\boldsymbol{\nu}}(s)
    -
    V_c^{\boldsymbol{\pi}}(s).
\end{align}
Therefore, by the one-step attack cost bound,
\begin{align}
    \delta_c^{\tilde{\boldsymbol{\pi}}}
    &=
    \max_s
    \left|
    \tilde{V}_c^{\boldsymbol{\pi},\boldsymbol{\nu}}(s)
    -
    V_c^{\boldsymbol{\pi}}(s)
    \right| \nonumber\\
    &\le
    2m^2\epsilon L_{max}c_{ind}
    \left(
    \max_s p_s+\frac{\gamma}{1-\gamma}
    \right).
\end{align}
Since the unattacked agents execute the same policies under
$\tilde{\boldsymbol{\pi}}$ and $\boldsymbol{\pi}$, the policy shift is entirely
induced by the attacked subset. Hence,
\begin{equation}
    \max_s
    D_{\mathrm{TV}}
    \left[
    \tilde{\boldsymbol{\pi}}(\cdot\mid s)
    \,\middle\|\,
    \boldsymbol{\pi}(\cdot\mid s)
    \right]
    \le
    mL_{max}\epsilon.
\end{equation}
Substituting these two bounds into Eq.~\eqref{eq:joint-cpo-bound}, we obtain
\begin{align}
    &V_c^{\tilde{\boldsymbol{\pi}}}(\mu_0)
    -
    V_c^{\boldsymbol{\pi}}(\mu_0) \nonumber\\
    &\le
    \delta_c^{\tilde{\boldsymbol{\pi}}}
    \left(
    \frac{1}{1-\gamma}
    +
    \frac{2\gamma mL_{max}\epsilon}{(1-\gamma)^2}
    \right) \nonumber\\
    &\le
    2m^2\epsilon L_{max}c_{ind}
    \left(
    \frac{1}{1-\gamma}
    +
    \frac{2\gamma m\epsilon L_{max}}{(1-\gamma)^2}
    \right)
    \left(
    \max_s p_s+\frac{\gamma}{1-\gamma}
    \right).
\end{align}
Finally, if the clean fixed joint policy is feasible before attack, i.e.,
\begin{equation}
    V_c^{\boldsymbol{\pi}}(\mu_0)\le \kappa,
\end{equation}
then
\begin{equation}
    V_c^{\tilde{\boldsymbol{\pi}}}(\mu_0)
    \le
    \kappa
    +
    2m^2\epsilon L_{max}c_{ind}
    \left(
    \frac{1}{1-\gamma}
    +
    \frac{2\gamma m\epsilon L_{max}}{(1-\gamma)^2}
    \right)
    \left(
    \max_s p_s+\frac{\gamma}{1-\gamma}
    \right).
\end{equation}

We next establish a lower bound on the episodic safety cost under attack. By definition, $V_c^{\tilde{\boldsymbol{\pi}}}(\mu_0)$ is the expected discounted cumulative safety cost under the attacked joint policy. Since the safety cost is non-negative and $\gamma\in[0,1)$, every discounted cost term is non-negative. Therefore,
\begin{equation}
V_c^{\tilde{\boldsymbol{\pi}}}(\mu_0)\ge 0.
\end{equation}

\subsection{Cost Bound under a $K$-Step Attack}

We now consider a finite-duration attack in which the adversary perturbs the
observations of the compromised agents during the first $K$ environment
timesteps and terminates the attack thereafter. The attacked subset
$\mathcal{M}$ remains fixed throughout the attack and satisfies
$|\mathcal{M}|=m$.

Define the resulting non-stationary joint policy
$\tilde{\boldsymbol{\pi}}^{(K)}
=
\{\tilde{\boldsymbol{\pi}}^{(K)}_t\}_{t=0}^{\infty}$
as
\begin{equation}
\tilde{\boldsymbol{\pi}}^{(K)}_t
(\boldsymbol{a}\mid s)
=
\begin{cases}
\tilde{\boldsymbol{\pi}}(\boldsymbol{a}\mid s),
& 0\leq t < K,\\[2mm]
\boldsymbol{\pi}(\boldsymbol{a}\mid s),
& t\geq K,
\end{cases}
\end{equation}
where, during each attacked timestep $t<K$,
\begin{equation}
\tilde{\boldsymbol{\pi}}(\boldsymbol{a}\mid s)
=
\left(
\prod_{i\in\mathcal{M}}
\pi^i(a^i\mid \tilde{s}^i)
\right)
\left(
\prod_{j\in\mathcal{U}}
\pi^j(a^j\mid s^j)
\right),
\qquad
\tilde{s}^i\in B_p^\epsilon(s^i).
\end{equation}

Hence, the agents execute the attacked joint policy during the first $K$
timesteps and revert to the clean joint policy $\boldsymbol{\pi}$ from
timestep $K$ onward.

Recall that the clean cost advantage is
\begin{equation}
A_c^{\boldsymbol{\pi}}(s,\boldsymbol{a})
=
Q_c^{\boldsymbol{\pi}}(s,\boldsymbol{a})
-
V_c^{\boldsymbol{\pi}}(s).
\end{equation}

For any attacked timestep $t<K$ and any state $s$, the expected clean
advantage under the attacked joint policy satisfies
\begin{align}
&\mathbb{E}_{\boldsymbol{a}\sim
\tilde{\boldsymbol{\pi}}(\cdot\mid s)}
\left[
A_c^{\boldsymbol{\pi}}(s,\boldsymbol{a})
\right]
\nonumber\\
&=
\mathbb{E}_{\boldsymbol{a}\sim
\tilde{\boldsymbol{\pi}}(\cdot\mid s)}
\left[
Q_c^{\boldsymbol{\pi}}(s,\boldsymbol{a})
-
V_c^{\boldsymbol{\pi}}(s)
\right]
\nonumber\\
&=
\tilde{V}_c^{\boldsymbol{\pi},\boldsymbol{\nu}}(s)
-
V_c^{\boldsymbol{\pi}}(s).
\end{align}

By the one-step attack cost bound established in Theorem~1,
\begin{align}
&\mathbb{E}_{\boldsymbol{a}\sim
\tilde{\boldsymbol{\pi}}(\cdot\mid s)}
\left[
A_c^{\boldsymbol{\pi}}(s,\boldsymbol{a})
\right]
\nonumber\\
&\leq
2m^2\epsilon L_{max}c_{ind}
\left(
p_s+\frac{\gamma}{1-\gamma}
\right)
.
\label{eq:k-step-advantage-bound}
\end{align}

We next express the total cost difference induced by the $K$-step attack.
Because the attacked policy reverts to the clean policy after timestep $K$,
the finite-duration performance difference identity gives
\begin{align}
&V_c^{\tilde{\boldsymbol{\pi}}^{(K)}}(\mu_0)
-
V_c^{\boldsymbol{\pi}}(\mu_0)
\nonumber\\
&=
\mathbb{E}_{\tau\sim
\tilde{\boldsymbol{\pi}}^{(K)}}
\left[
\sum_{t=0}^{K-1}
\gamma^t
A_c^{\boldsymbol{\pi}}
(s_t,\boldsymbol{a}_t)
\right].
\label{eq:k-step-performance-difference}
\end{align}

For completeness, Eq.~\eqref{eq:k-step-performance-difference} follows from
the Bellman definition
\begin{equation}
Q_c^{\boldsymbol{\pi}}(s_t,\boldsymbol{a}_t)
=
\mathbb{E}
\left[
c(s_t,\boldsymbol{a}_t,s_{t+1})
+
\gamma V_c^{\boldsymbol{\pi}}(s_{t+1})
\right].
\end{equation}
Therefore,
\begin{align}
&\mathbb{E}_{\tau\sim
\tilde{\boldsymbol{\pi}}^{(K)}}
\left[
\sum_{t=0}^{K-1}
\gamma^t
A_c^{\boldsymbol{\pi}}
(s_t,\boldsymbol{a}_t)
\right]
\nonumber\\
&=
\mathbb{E}_{\tau\sim
\tilde{\boldsymbol{\pi}}^{(K)}}
\left[
\sum_{t=0}^{K-1}
\gamma^t
c(s_t,\boldsymbol{a}_t,s_{t+1})
+
\gamma^K V_c^{\boldsymbol{\pi}}(s_K)
-
V_c^{\boldsymbol{\pi}}(s_0)
\right].
\end{align}
Since the agents follow $\boldsymbol{\pi}$ from timestep $K$ onward,
$\gamma^K V_c^{\boldsymbol{\pi}}(s_K)$ is exactly the expected discounted
tail cost after the attack terminates. Consequently, the right-hand side is
equal to
\begin{equation}
V_c^{\tilde{\boldsymbol{\pi}}^{(K)}}(\mu_0)
-
V_c^{\boldsymbol{\pi}}(\mu_0).
\end{equation}

Applying the per-timestep advantage bound, we obtain
\begin{align}
&V_c^{\tilde{\boldsymbol{\pi}}^{(K)}}(\mu_0)
-
V_c^{\boldsymbol{\pi}}(\mu_0)
\nonumber\\
&\leq
\sum_{t=0}^{K-1}
\gamma^t
\left[
2m^2\epsilon L_{max}c_{ind}
\left(
\max_s p_s+\frac{\gamma}{1-\gamma}
\right)
\right]
\nonumber\\
&=
2m^2\epsilon L_{max}c_{ind}
\left(
\max_s p_s+\frac{\gamma}{1-\gamma}
\right)
\sum_{t=0}^{K-1}\gamma^t.
\end{align}

Using the finite geometric series
\begin{equation}
\sum_{t=0}^{K-1}\gamma^t
=
\frac{1-\gamma^K}{1-\gamma},
\end{equation}
the cost increase caused by attacking $m$ agents for $K$ timesteps is bounded
by
\begin{align}
&V_c^{\tilde{\boldsymbol{\pi}}^{(K)}}(\mu_0)
-
V_c^{\boldsymbol{\pi}}(\mu_0)
\nonumber\\
&\leq
2m^2\epsilon L_{max}c_{ind}
\left(
\max_s p_s+\frac{\gamma}{1-\gamma}
\right)
\frac{1-\gamma^K}{1-\gamma}.
\label{eq:k-step-cost-bound}
\end{align}

The cost value under the $K$-step attack satisfies
\begin{equation}
 V_c^{\tilde{\boldsymbol{\pi}}^{(K)}}(\mu_0)
\leq
\kappa
+
2m^2\epsilon L_{max}c_{ind}
\left(
\max_s p_s+\frac{\gamma}{1-\gamma}
\right)
\frac{1-\gamma^K}{1-\gamma}.
\end{equation}

As $K\rightarrow\infty$,
\begin{equation}
\frac{1-\gamma^K}{1-\gamma}
\longrightarrow
\frac{1}{1-\gamma},
\end{equation}
which recovers the corresponding persistent-attack accumulation factor.

For $K\ll 1/(1-\gamma)$ and $\gamma$ close to one,
\begin{equation}
\frac{1-\gamma^K}{1-\gamma}
\approx K,
\end{equation}
and the finite-duration attack cost bound therefore scales approximately as
\begin{equation}
\mathcal{O}\left(Km^2\epsilon\right).
\end{equation}

\section{Proof of Joint Attack Dominance}
\label{appB}
We show that jointly optimizing the attacked actions of multiple victim agents is no worse than independently optimizing each victim attack when the safety cost critic exhibits non-separable coupling across agents. Moreover, when the attack induces positive non-separable safety coupling, the joint optimization attack is strictly better.

Let $\boldsymbol{a}_{\mathcal{M}}=\{a^i\}_{i\in\mathcal{M}}$ be a candidate
attacked joint action of the victim agents, let
$\boldsymbol{a}_{\mathcal{M}}^0$ be their
nominal clean joint action, and let $\boldsymbol{a}_{\mathcal{U}}$ denote the
actions of the benign agents. For a fixed state $s$, the true joint safety cost
increase caused by changing all victim agents from
$\boldsymbol{a}_{\mathcal{M}}^0$ to $\boldsymbol{a}_{\mathcal{M}}$ is defined
as
\begin{equation}
    \Delta_{\mathrm{joint}}
    (s;\boldsymbol{a}_{\mathcal{M}},\boldsymbol{a}_{\mathcal{U}})
    =
    Q_c^{\boldsymbol{\pi}}
    (s,\boldsymbol{a}_{\mathcal{M}},\boldsymbol{a}_{\mathcal{U}})
    -
    Q_c^{\boldsymbol{\pi}}
    (s,\boldsymbol{a}_{\mathcal{M}}^0,\boldsymbol{a}_{\mathcal{U}}).
\end{equation}
For each victim agent $i\in\mathcal{M}$, the corresponding single-victim
safety cost increase is defined by changing only agent $i$ while keeping all
other victim agents at their nominal clean actions:
\begin{equation}
    \Delta_i(s;a^i,\boldsymbol{a}_{\mathcal{U}})
    =
    Q_c^{\boldsymbol{\pi}}
    (s,a^i,\boldsymbol{a}_{\mathcal{M}\setminus\{i\}}^0,
    \boldsymbol{a}_{\mathcal{U}})
    -
    Q_c^{\boldsymbol{\pi}}
    (s,\boldsymbol{a}_{\mathcal{M}}^0,\boldsymbol{a}_{\mathcal{U}}).
\end{equation}
The additive independent approximation of the multi-victim attack effect is
then
\begin{equation}
    \Delta_{\mathrm{add}}
    (s;\boldsymbol{a}_{\mathcal{M}},\boldsymbol{a}_{\mathcal{U}})
    =
    \sum_{i\in\mathcal{M}}
    \Delta_i(s;a^i,\boldsymbol{a}_{\mathcal{U}}).
\end{equation}
We define non-separable safety coupling as the residual between the true joint
safety cost increase and this additive approximation:
\begin{align}
    \Delta_{\mathrm{NSSC}}
    (s;\boldsymbol{a}_{\mathcal{M}},\boldsymbol{a}_{\mathcal{U}})
    &:=
    \Delta_{\mathrm{joint}}
    (s;\boldsymbol{a}_{\mathcal{M}},\boldsymbol{a}_{\mathcal{U}})
    -
    \Delta_{\mathrm{add}}
    (s;\boldsymbol{a}_{\mathcal{M}},\boldsymbol{a}_{\mathcal{U}}) \nonumber\\
    &=
    \left[
    Q_c^{\boldsymbol{\pi}}
    (s,\boldsymbol{a}_{\mathcal{M}},\boldsymbol{a}_{\mathcal{U}})
    -
    Q_c^{\boldsymbol{\pi}}
    (s,\boldsymbol{a}_{\mathcal{M}}^0,\boldsymbol{a}_{\mathcal{U}})
    \right] \nonumber\\
    &\quad
    -
    \sum_{i\in\mathcal{M}}
    \left[
    Q_c^{\boldsymbol{\pi}}
    (s,a^i,\boldsymbol{a}_{\mathcal{M}\setminus\{i\}}^0,
    \boldsymbol{a}_{\mathcal{U}})
    -
    Q_c^{\boldsymbol{\pi}}
    (s,\boldsymbol{a}_{\mathcal{M}}^0,\boldsymbol{a}_{\mathcal{U}})
    \right].
\end{align}

Let $\mathcal{A}_i$ denote the action space of victim agent $i$, and let
\begin{equation}
    \mathcal{A}_{\mathcal{M}}
    =
    \prod_{i\in\mathcal{M}}\mathcal{A}_i
\end{equation}
be the joint target-action space of the compromised agents. The independent attack optimizes
each victim action separately according to its single-victim marginal
safety cost increase:
\begin{equation}
    a_{\mathrm{ind}}^i
    \in
    \arg\max_{a^i\in\mathcal{A}_i}
    \Delta_i(s;a^i,\boldsymbol{a}_{\mathcal{U}}),
    \qquad i\in\mathcal{M}.
\end{equation}
Let
$\boldsymbol{a}_{\mathcal{M}}^{\mathrm{ind}}
=
\{a_{\mathrm{ind}}^i\}_{i\in\mathcal{M}}$
be the joint action obtained by combining these independently optimized
victim actions. The corresponding independent additive attack value is
\begin{equation}
    J_{\mathrm{ind}}^\star(s,\boldsymbol{a}_{\mathcal{U}})
    =
    \sum_{i\in\mathcal{M}}
    \Delta_i(s;a_{\mathrm{ind}}^i,\boldsymbol{a}_{\mathcal{U}}).
\end{equation}
In contrast, JOMAS directly optimizes the true joint safety cost increase:
\begin{equation}
    J_{\mathrm{joint}}^\star(s,\boldsymbol{a}_{\mathcal{U}})
    =
    \max_{\boldsymbol{a}_{\mathcal{M}}\in
    \mathcal{A}_{\mathcal M}}
    \Delta_{\mathrm{joint}}
    (s;\boldsymbol{a}_{\mathcal{M}},\boldsymbol{a}_{\mathcal{U}}).
\end{equation}
Let $\boldsymbol{a}_{\mathcal{M}}^{\mathrm{ind}}$ be the joint action obtained by independently optimizing each victim attack. Using the definition of NSSC, this objective can be rewritten as
\begin{align}
    J_{\mathrm{joint}}^\star(s,\boldsymbol{a}_{\mathcal{U}})
    &=
    \max_{\boldsymbol{a}_{\mathcal{M}}\in
    \mathcal{A}_{\mathcal M}}
    \left[
    \sum_{i\in\mathcal{M}}
    \Delta_i(s;a^i,\boldsymbol{a}_{\mathcal{U}})
    +
    \Delta_{\mathrm{NSSC}}
    (s;\boldsymbol{a}_{\mathcal{M}},\boldsymbol{a}_{\mathcal{U}})
    \right]  \\
    &\ge
    J_{\mathrm{ind}}^\star(s,\boldsymbol{a}_{\mathcal{U}})
    +
    \Delta_{\mathrm{NSSC}}
    (s;\boldsymbol{a}_{\mathcal{M}}^{\mathrm{ind}},
    \boldsymbol{a}_{\mathcal{U}}).
\end{align}
Consequently, if the independently combined attack induces non-negative NSSC, if
\begin{equation}
    \Delta_{\mathrm{NSSC}}
    (s;\boldsymbol{a}_{\mathcal{M}}^{\mathrm{ind}},
    \boldsymbol{a}_{\mathcal{U}})
    >0,
\end{equation}
then
\begin{equation}
    J_{\mathrm{joint}}^\star(s,\boldsymbol{a}_{\mathcal{U}})
    >
    J_{\mathrm{ind}}^\star(s,\boldsymbol{a}_{\mathcal{U}}).
\end{equation}

\section{Constraint-Induced Cooperative Resistance}
\label{appNew}
We next characterize why cooperative resistance naturally arises in constrained cooperative control like SMARL. Consider a fixed state $s$ and write the
joint action as $\mathbf{a}=(\mathbf{a}_{\mathcal M},
\mathbf{a}_{\mathcal U})$, where $\mathcal M$ and $\mathcal U$ denote the
compromised and benign agent sets, respectively. Let
$Q_R^{\boldsymbol{\pi}}(s,\mathbf{a})$ and
$Q_c^{\boldsymbol{\pi}}(s,\mathbf{a})$ denote the reward and safety cost
critics under the clean joint policy $\boldsymbol{\pi}$.

For a fixed action $\mathbf{a}_{\mathcal M}$ of the compromised agents,
consider the local constrained best response of the benign agents
\begin{equation}
\mathbf{a}_{\mathcal U}^{\star}(\mathbf{a}_{\mathcal M})
\in
\arg\max_{\mathbf{a}_{\mathcal U}}
Q_R^{\boldsymbol{\pi}}
(s,\mathbf{a}_{\mathcal M},\mathbf{a}_{\mathcal U})
\quad
\text{s.t.}\quad
Q_c^{\boldsymbol{\pi}}
(s,\mathbf{a}_{\mathcal M},\mathbf{a}_{\mathcal U})
\leq \bar c_s ,
\label{eq:benign_best_response}
\end{equation}
where $\bar c_s$ denotes the local safety boundary induced by an active
safety constraint.

Let
$\mathbf{a}^{0}
=(\mathbf{a}_{\mathcal M}^{0},\mathbf{a}_{\mathcal U}^{0})$
be a locally optimal clean joint action satisfying
\begin{equation}
Q_c^{\boldsymbol{\pi}}(s,\mathbf{a}^{0})=\bar c_s .
\end{equation}

Assume that $Q_R^{\boldsymbol{\pi}}$ and $Q_c^{\boldsymbol{\pi}}$ are
continuously differentiable, the active safety constraint is regular and
locally stable under sufficiently small perturbations of
$\mathbf{a}_{\mathcal M}$, and the benign agents have nonzero control
authority over the safety cost,
\begin{equation}
\nabla_{\mathbf{a}_{\mathcal U}}
Q_c^{\boldsymbol{\pi}}(s,\mathbf{a}^{0})
\neq \mathbf{0}.
\end{equation}
Consider an infinitesimal deviation $\mathbf d_{\mathcal M}$ of the
compromised agents that increases the safety cost,
\begin{equation}
\Delta_C^{\mathcal M}
=
\nabla_{\mathbf{a}_{\mathcal M}}
Q_c^{\boldsymbol{\pi}}(s,\mathbf{a}^{0})^\top
\mathbf d_{\mathcal M}
>0.
\label{eq:attack_cost_direction}
\end{equation}
Then any differentiable locally feasible constrained response of the
benign agents contains a nonzero cost-reducing component
$\mathbf d_{\mathcal U}$ satisfying
\begin{equation}
\nabla_{\mathbf{a}_{\mathcal U}}
Q_c^{\boldsymbol{\pi}}(s,\mathbf{a}^{0})^\top
\mathbf d_{\mathcal U}
=
-\Delta_C^{\mathcal M}
<0.
\label{eq:cooperative_resistance_direction}
\end{equation}
Hence, a cost-increasing deviation of the compromised agents necessarily
induces a compensatory safety direction from the benign agents whenever
the active safety constraint is coupled across the two agent subsets.

Define a perturbed compromised action
\begin{equation}
\mathbf{a}_{\mathcal M}(\tau)
=
\mathbf{a}_{\mathcal M}^{0}
+
\tau \mathbf d_{\mathcal M},
\end{equation}
for sufficiently small $\tau>0$, and let
$\mathbf{a}_{\mathcal U}^{\star}(\tau)$ denote the corresponding local
solution of Eq.~\eqref{eq:benign_best_response}. Since the safety constraint is active at $\tau=0$ and remains active locally under the local active-set stability assumption,
\begin{equation}
Q_c^{\boldsymbol{\pi}}
\left(
s,
\mathbf{a}_{\mathcal M}(\tau),
\mathbf{a}_{\mathcal U}^{\star}(\tau)
\right)
=
\bar c_s .
\label{eq:active_constraint_path}
\end{equation}

Differentiating Eq.~\eqref{eq:active_constraint_path} with respect to
$\tau$ at $\tau=0$ gives
\begin{align}
&
\nabla_{\mathbf{a}_{\mathcal M}}
Q_c^{\boldsymbol{\pi}}(s,\mathbf{a}^{0})^\top
\mathbf d_{\mathcal M}
\nonumber\\
&\quad+
\nabla_{\mathbf{a}_{\mathcal U}}
Q_c^{\boldsymbol{\pi}}(s,\mathbf{a}^{0})^\top
\left.
\frac{d\mathbf{a}_{\mathcal U}^{\star}(\tau)}
{d\tau}
\right|_{\tau=0}
=0.
\end{align}
Let
\begin{equation}
\mathbf d_{\mathcal U}
=
\left.
\frac{d\mathbf{a}_{\mathcal U}^{\star}(\tau)}
{d\tau}
\right|_{\tau=0}.
\end{equation}
Using Eq.~\eqref{eq:attack_cost_direction}, we obtain
\begin{equation}
\nabla_{\mathbf{a}_{\mathcal U}}
Q_c^{\boldsymbol{\pi}}(s,\mathbf{a}^{0})^\top
\mathbf d_{\mathcal U}
=
-
\nabla_{\mathbf{a}_{\mathcal M}}
Q_c^{\boldsymbol{\pi}}(s,\mathbf{a}^{0})^\top
\mathbf d_{\mathcal M}
=
-\Delta_C^{\mathcal M}
<0.
\end{equation}
Therefore, the benign response necessarily moves in a direction that
reduces the safety cost. Since $\Delta_C^{\mathcal M}>0$, this response
cannot be zero. We refer to this safety-restoring adjustment as \emph{Cooperative Resistance}.
 
For a finite but sufficiently small deviation, let
$\boldsymbol{\nu}_{\mathcal U}$ denote the corresponding benign action
change. A first-order expansion gives
\begin{equation}
\nabla_{\mathbf{a}_{\mathcal U}}
Q_c^{\boldsymbol{\pi}}(s,\mathbf{a}^{0})^\top
\boldsymbol{\nu}_{\mathcal U}
=
-
\nabla_{\mathbf{a}_{\mathcal M}}
Q_c^{\boldsymbol{\pi}}(s,\mathbf{a}^{0})^\top
\Delta\mathbf{a}_{\mathcal M}
+
o(\|\Delta\mathbf{a}_{\mathcal M}\|),
\end{equation}
which provides the local theoretical basis for the Cooperative Resistance
term.

Moreover, if the compromised deviation is simultaneously reward-increasing,
\begin{equation}
\nabla_{\mathbf{a}_{\mathcal M}}
Q_R^{\boldsymbol{\pi}}(s,\mathbf{a}^{0})^\top
\mathbf d_{\mathcal M}>0,
\end{equation}
and this reward gain dominates the reward loss induced by the benign
compensation,
\begin{equation}
\nabla_{\mathbf{a}_{\mathcal M}}
Q_R^{\boldsymbol{\pi}}(s,\mathbf{a}^{0})^\top
\mathbf d_{\mathcal M}
+
\nabla_{\mathbf{a}_{\mathcal U}}
Q_R^{\boldsymbol{\pi}}(s,\mathbf{a}^{0})^\top
\mathbf d_{\mathcal U}
\geq 0,
\end{equation}
then the joint system can preserve or improve task reward while the benign
agents absorb the safety correction.  

\section{Victim SMARL Algorithm}
\label{appC}
This section reports the implementation and training details for the victim safe multi-agent reinforcement learning algorithms. All victim policies are trained under centralized training and decentralized execution. All experiments were implemented in PyTorch and conducted on NVIDIA A10 and NVIDIA RTX 4090 GPUs. The hyperparameters of MinMax Attack are typically set to $K_a=5$, $K_\nu=10$, and $K_\delta=10$.

To follow the standard notation of the victim algorithms, within
this section, $\alpha$ denotes the entropy coefficient of WCMASAC and HASAC-Lag. 

\subsection{MAPPO-Lag}
MAPPO-Lagrangian is an on-policy safe multi-agent actor-critic algorithm that augments
MAPPO with Lagrangian relaxation for constrained policy optimization \cite{26_gu2023safe}. During each update,
agents are optimized sequentially, and the safety constraint is incorporated into the actor
objective through non-negative Lagrange multipliers. Let
\[
\rho_i^t(\theta_i)
=
\frac{\pi_{\theta_i}(a_t^i \mid s_t^i)}
{\pi_{\theta_i^{\mathrm{old}}}(a_t^i \mid s_t^i)}
\]
denote the policy probability ratio of agent $i$. Let
$\widehat A_i^{r}(s_t,\mathbf a_t)$ denote the reward advantage used for the sequential
multi-agent policy update, and let
$\widehat A_{i}^{c}(s_t,a_t^i)$ denote the estimated cost advantage of agent $i$. The Lagrangian-adjusted advantage is defined as
\begin{equation}
\widehat A_i^{\lambda}(s_t,\mathbf a_t)
=
\widehat A_i^{r}(s_t,\mathbf a_t)
- \lambda_{i}
\widehat A_{i}^{c}(s_t,a_t^i),
\end{equation}
where $\lambda_i\geq 0$ is the Lagrange multiplier associated with the
safety constraint of agent $i$. The actor of agent $i$ is then
updated by maximizing the clipped surrogate objective
\begin{equation}
\begin{aligned}
&\mathcal{J}^{\mathrm{MAPPO\text{-}Lag}}_i(\theta_i)\\
&=\mathbb{E}_t\!\left[\min\!\left\{
\begin{aligned}
&\rho_i^t(\theta_i)\widehat A_i^{\lambda}(s_t,\mathbf a_t),\\
&\operatorname{clip}\!\left(
\rho_i^t(\theta_i),
1-\epsilon_{\mathrm{clip}},
1+\epsilon_{\mathrm{clip}}
\right)
\widehat A_i^{\lambda}(s_t,\mathbf a_t)
\end{aligned}
\right\}\right].
\end{aligned}
\end{equation}
Here, $\epsilon_{\mathrm{clip}}$ denotes the PPO clipping parameter. The multiplier is updated by projected gradient ascent according to the empirical constraint
violation, thereby penalizing policies whose expected safety cost exceeds the prescribed safety
threshold.

\subsection{MACPO}
Multi-Agent Constrained Policy Optimization (MACPO) is an on-policy safe MARL algorithm
based on constrained policy optimization~\cite{26_gu2023safe}. MACPO updates agents
sequentially and solves a local constrained optimization problem for each agent. For agent $i$,
let $\theta_i$ denote its policy parameters and let $\theta_i^k$ denote the parameters before the
$k$-th update. The local update is
\begin{equation}
\begin{aligned}
\theta_i^{k+1}
=
\arg\max_{\theta_i}
\quad &
g_i^\top(\theta_i-\theta_i^k)
\\
\mathrm{s.t.}
\quad &
c_i + b_i^\top(\theta_i-\theta_i^k) \leq 0,
\\
&
\frac{1}{2}(\theta_i-\theta_i^k)^\top H_i(\theta_i-\theta_i^k)
\leq \delta .
\end{aligned}
\end{equation}
Here, $g_i$ is the gradient of the reward objective with respect to $\theta_i$, $b_i$ is the
gradient of the cost constraint with respect to $\theta_i$, and $c_i$ is the current constraint
violation, i.e., the estimated cost minus the safety threshold. $H_i$ is the Hessian of the
average KL divergence between the old and updated policy of agent $i$, and $\delta$ is the
trust-region radius. The last constraint is the second-order approximation of the KL trust region.

\subsection{WCMASAC}
Worst-Case Multi-Agent Soft Actor-Critic (WCMASAC) is an off-policy safe MARL algorithm that combines maximum-entropy optimization with a distributional safety critic \cite{46_liu2026safe}. Let $\mathcal{N}=\{1,\ldots,n\}$ denote the set of agents, $s_t$ the state, $a_t^i$ the action of agent $i$, and $\mathbf a_t=(a_t^1,\ldots,a_t^n)$ the joint action.
The joint policy is factorized as $\pi(\mathbf a_t|s_t)=\prod_{i=1}^n\pi_i(a_t^i|s_t^i)$. WCMASAC optimizes
\begin{equation}
\begin{aligned}
\max_{\pi}\quad
&\mathbb{E}_{\pi}
\left[
\sum_{t=0}^{\infty}
\gamma^t
\left(
R(s_t,\mathbf a_t)
+
\alpha\sum_{i=1}^{n}\mathcal{H}(\pi_i(\cdot|s_t))
\right)
\right]
\\
\mathrm{s.t.}\quad
&
\phi^{\pi}_{1-\omega}
\left(
C_{i}(s_t,a_t^i,\mathbf a_t^{-i})
\right)
\leq d_{i},
\qquad
i\in\mathcal{N}, 
\end{aligned}
\end{equation}
Here, $R(s_t,\mathbf a_t)$ is the team reward, $\gamma$ is the discount factor, $\alpha$ is the entropy temperature, and $\mathcal{H}(\cdot)$ denotes policy entropy. The notation $\mathbf a_t^{-i}$ denotes the actions of all agents except $i$. The function $C_{i}$ is the safety cost associated with agent $i$, $\phi^{\pi}_{1-\omega}(\cdot)$ is the CVaR risk measure, $\omega$ is the risk level, and $d_{i}$ is the corresponding safety threshold.

For implementation, WCMASAC uses a centralized soft reward critic $Q_{\theta_r}$, a distributional cost critic $C_{\theta_{c}}$, and sequentially updated actors. Let $i_{1:n}$ be a sampled update order and let $i_k$ be the $k$-th updated agent. Define $a_{<k}=a_{i_{1:k-1}}$ as the actions of agents updated before $i_k$, and
$\tilde{\mathbf a}_k=(a_{<k},a_{i_k})$. We use the compact expectation
\begin{equation}
\mathbb{E}_k[\cdot]
=
\mathbb{E}_{(s,\mathbf a)\sim\mathcal D}
\mathbb{E}_{a_{<k}\sim\pi_{\hat{\theta}_a^{i_{1:k-1}}},
a_{i_k}\sim\pi_{\theta_a^{i_k}}}
[\cdot],
\end{equation}
where $\mathcal D$ is the replay buffer, $\theta_a^{i_k}$ is the actor parameter of agent $i_k$, and $\hat{\theta}_a^{i_{1:k-1}}$ denotes the updated actor parameters of the preceding agents in the current update order. The actor loss for agent $i_k$ is
\begin{equation}
\begin{aligned}
\mathcal{L}^{i_k}_{\pi}(\theta_a^{i_k})
=
-\mathbb{E}_k
\Big[
&Q_{\theta_r}(s,\tilde{\mathbf a}_k)
-
\lambda_{i_k}
\phi_{1-\omega}
\big(
C_{\theta_{c}}(s,\tilde{\mathbf a}_k)
\big)
\\
&-
\alpha
\log \pi_{\theta_a^{i_k}}(a_{i_k}|s^{i_k})
\Big].
\end{aligned}
\end{equation}
Here, $\theta_r$ and $\theta_{c}$ are the parameters of the reward critic and the cost critic, respectively. The scalar $\lambda_{i_k}\geq 0$ is the safety multiplier for agent
$i_k$, and $\alpha$ is the entropy coefficient. The term $\phi_{1-\omega}(C_{\theta_{c}})$ penalizes the worst-case tail risk of the learned cost distribution.

\subsection{HASAC-Lag}
HASAC-Lag is the Lagrangian expected-cost variant of Heterogeneous-Agent Soft Actor-Critic used in our experiments \cite{46_liu2026safe}. The base HASAC algorithm is an off-policy maximum-entropy MARL method with a centralized soft reward critic and decentralized stochastic actors \cite{47_liu2023maximum}. HASAC-Lag adds an expected-cost penalty to the HASAC-Lag actor objective. Using the same sequential-update notation as above, let $Q_{\theta_r}$ denote the centralized soft reward critic and $Q_{\theta_c}$ denote the centralized expected-cost critic.
The actor loss for agent $i_k$ is
\begin{equation}
\begin{aligned}
\mathcal{L}^{i_k}_{\pi}(\theta_a^{i_k})
=
-\mathbb{E}_k
\Big[
&Q_{\theta_r}(s,\tilde{\mathbf a}_k)
-
\lambda
Q_{\theta_c}(s,\tilde{\mathbf a}_k)
\\
&-
\alpha
\log \pi_{\theta_a^{i_k}}(a_{i_k}|s^{i_k})
\Big].
\end{aligned}
\end{equation}
Here, $\mathbb{E}_k[\cdot]$ is the replay-buffer and sequential-action expectation defined above, $\theta_a^{i_k}$ is the actor parameter of agent $i_k$, $\theta_r$ and $\theta_c$ are the reward-critic and cost-critic parameters, $\lambda\geq 0$ is the Lagrange multiplier, and $\alpha$ is the entropy coefficient. Unlike WCMASAC, HASAC-Lag uses the expected-cost critic $Q_{\theta_c}$ rather than a CVaR distributional cost critic.

The Lagrange multiplier is updated by
\begin{equation}
\lambda
\leftarrow
\left[
\lambda
+
\eta_{\lambda}
\left(
\widehat J_c(\pi)-d
\right)
\right]_+ ,
\end{equation}
where $\eta_{\lambda}$ is the multiplier learning rate, $\widehat J_c(\pi)$ is the empirical
episode-level cost under policy $\pi$, $d$ is the safety threshold, and
$[x]_+=\max(x,0)$ projects the multiplier onto the non-negative domain.
\subsection{Training Hyperparameters}
The hyperparameters used for four victim algorithms are summarised in Table~\ref{tabapd}. Parameters that are not used by a specific algorithm are marked as N/A. We follow the reference configuration in the safe MARL benchmark \cite{26_gu2023safe,46_liu2026safe,47_liu2023maximum}.

\begin{table}[ht]
\centering
\caption{Training hyperparameters of victim safe MARL algorithms.}
\label{tabapd}
\small
\setlength{\tabcolsep}{4pt}
\renewcommand{\arraystretch}{1.0}
\resizebox{\columnwidth}{!}{%
\begin{tabular}{lcccc}
\toprule
\textbf{Hyperparameter} 
& \textbf{MAPPO-Lag} 
& \textbf{MACPO} 
& \textbf{WCMASAC} 
& \textbf{HASAC-Lag} \\
\midrule
Training paradigm 
& On-policy 
& On-policy 
& Off-policy 
& Off-policy \\

Actor learning rate 
& $9\times 10^{-5}$ 
& $5\times 10^{-5}$  
& $1\times 10^{-3}$ 
& $1\times 10^{-3}$ \\

Critic learning rate 
& $5\times 10^{-3}$ 
& $5\times 10^{-3}$ 
& $1\times 10^{-3}$ 
& $1\times 10^{-3}$ \\

Discount factor $\gamma$ 
& $0.99$ 
& $0.96$ 
& $0.99$ 
& $0.99$ \\

Batch size 
& $16000$ 
& $16000$ 
& $1000$ 
& $1000$ \\

Rollout threads 
& $16$ 
& $16$ 
& $10$ 
& $10$ \\

Optimizer 
& Adam 
& Adam 
& Adam 
& Adam \\

Entropy coefficient $\alpha$ 
& N/A 
& N/A 
& Auto-tuned 
& Auto-tuned \\

Lagrangian learning rate 
& $1\times 10^{-3}$ 
& N/A 
& $1\times 10^{-3}$ 
& $1\times 10^{-3}$ \\

PPO clip parameter 
& $0.2$ 
& N/A 
& N/A 
& N/A \\

KL threshold 
& N/A 
& $0.0065$ 
& N/A 
& N/A \\

Train interval 
& N/A 
& N/A 
& $50$ 
& $50$ \\
\bottomrule
\end{tabular}%
}
\end{table}

\section{Simulation Environment}
\label{appD}

This section describes the simulation environments used to evaluate the proposed attack. All environments involve continuous control and cooperative multi-agent decision making under safety constraints.

\subsection{Safe Multi-Agent MuJoCo}
Safe Multi-Agent MuJoCo is a safety-augmented extension of Multi-Agent MuJoCo \cite{51_todorov2012mujoco, 26_gu2023safe}. It preserves the original
multi-agent morphology, MuJoCo physics simulator, agent decomposition, and task reward, while introducing
explicit safety hazards and associated cost signals. In the HalfCheetah task, multiple agents jointly control
different body parts of a HalfCheetah robot and must coordinate their actions to achieve forward locomotion.
Compared with the standard Multi-Agent MuJoCo setting, the safe variant places the robot inside a constrained
corridor with dynamic obstacles. The corridor itself restricts the feasible movement region but does not directly
induce safety cost. Instead, safety violations are generated by moving pitfalls or obstacles: when the robot becomes
too close to an obstacle, the environment emits a unit cost.

Formally, let $y_{\mathrm{torso}, t+1}$ denote the $y$-coordinate of the robot torso at time $t+1$, and let
$y_{\mathrm{obstacle}}$ denote the $y$-coordinate of the moving obstacle. The instantaneous safety cost is defined as
\begin{equation}
c_t =
\begin{cases}
0, & \| y_{\mathrm{torso}, t+1} - y_{\mathrm{obstacle}} \|_2 \geq 9, \\
1, & \text{otherwise}.
\end{cases}
\end{equation}
This design requires agents not only to optimize locomotion reward but also to coordinate safely under dynamic
environmental hazards. Therefore, Safe Multi-Agent MuJoCo HalfCheetah provides a suitable benchmark for evaluating
whether localized observation perturbations can induce safety violations in cooperative locomotion.

\subsection{Safe Multi-Agent Isaac Gym}
\label{app:safe-ma-isaac-gym}

Safe Multi-Agent Isaac Gym (Safe MAIG) is a safety-augmented multi-agent robotic control benchmark built on top of Isaac Gym, a high-performance GPU-based simulation platform for robotic learning \cite{50_makoviychuk2021isaac}. Isaac Gym enables physics simulation and policy learning to be executed directly on GPU, which makes it suitable for large-scale dexterous manipulation experiments with parallel environments. In Safe MAIG, different body parts of the robotic hands are controlled by different agents, and the agents must
coordinate their actions to complete the manipulation objective while satisfying explicit safety constraints.

We use the ShadowHandReOrientation task as the representative high-dimensional dexterous manipulation environment. In this task, two Shadow Hands cooperatively manipulate two objects, and the objective is to reorient the objects around each other toward the desired configuration. The task requires fine-grained coordination across
multiple agents because the motion of each controlled finger or hand component affects the global manipulation outcome through coupled contact dynamics.  This task presents substantially higher-dimensional observations, more complex physical contacts, and stronger inter-agent dependency, making it a challenging benchmark for evaluating safety degradation under adversarial observation perturbations.

Following the Safe MAIG benchmark, the safety constraint is imposed on the range of motion of a constrained finger \cite{26_gu2023safe}. Let $F_{a_4,t+1}$ denote the motion degree of the fourth finger of the first hand at time $t+1$. The instantaneous safety cost is defined as
\begin{equation}
c_t =
\begin{cases}
1.0, & |F_{a_4,t+1}| \geq 0.1, \\
0, & \text{otherwise}.
\end{cases}
\end{equation}
Thus, a safety violation occurs when the constrained finger exceeds the prescribed motion range. This cost design allows us to evaluate whether an attack can induce constraint violations in a contact-rich cooperative manipulation task, even when the
underlying victim policies are trained to balance task performance and safety.

\subsection{Safe Highway-env}
 
We further evaluate our attack framework in a multi-agent autonomous driving environment
based on Highway-env. The task models an unsignalized four-way intersection, where multiple
vehicles approach the junction from different directions and must safely pass through the
intersection without traffic-light guidance. Each vehicle is controlled by an independent agent
and selects its driving action from local observations. The environment therefore requires
decentralized coordination, since unsafe acceleration, delayed yielding, or aggressive crossing
behaviors may increase the risk of near-collision among interacting vehicles.

The safety signal is defined through a time-to-collision (TTC) based cost. Importantly, TTC is
only considered for vehicles that are spatially close enough to interact. Let $d_{ij,t}$ denote the
distance between vehicles $i$ and $j$ at time $t$, and let $\mathrm{TTC}_{ij,t}$ denote their pairwise
time-to-collision under the current velocities. Given a proximity threshold $d_{\mathrm{safe}}$ and
a TTC threshold $\tau_{\mathrm{safe}}$, the instantaneous safety cost is defined as
\begin{equation}
c_t =
\alpha_{\mathrm{ttc}}
\sum_{i<j}
\mathbb{I}\!\left[d_{ij,t} \leq d_{\mathrm{safe}}\right]
\max\left(0, \tau_{\mathrm{safe}} - \mathrm{TTC}_{ij,t}\right),
\end{equation}
where $\alpha_{\mathrm{ttc}} > 0$ is a small scaling coefficient used to keep the safety penalty on a
comparable scale with the task reward. Under this definition, the cost is zero when vehicles are
sufficiently far apart.

This intersection environment complements the robotic control benchmarks by evaluating safety under structured traffic interactions. Unlike locomotion and dexterous manipulation tasks, this domain emphasizes decentralized driving, collision avoidance, and implicit negotiation among multiple autonomous vehicles. 

\section{Observation Attack Algorithms}
\label{appE}
We compare the proposed MinMax Attack against four representative baselines: Random perturbation, Maximal Action Difference (MAD) attack~\cite{48_zhang2020robust}, Max-Cost (MC) attack~\cite{45_liu2022robustness}, and Max-Reward (MR) attack~\cite{45_liu2022robustness}. Random perturbation serves as an unstructured noise baseline. MAD maximizes the deviation between perturbed and nominal actions. MC and MR use the victim critics to generate observation perturbations that maximize predicted safety cost or reward, respectively.

\subsection{Random Perturbation}
\label{appE1}
Random perturbation does not use victim-policy or critic information. For each compromised agent $i\in\mathcal{M}$, the perturbation is sampled independently from a uniform distribution:
\begin{equation}
    \delta_{i,t}
    \sim
    \mathrm{Uniform}(-\epsilon,\epsilon).
\end{equation}
The perturbed observation is then clipped to the valid observation range when such bounds are available. This baseline measures whether unstructured observation noise is sufficient to induce safety violations.


\subsection{Maximal Action Difference Attack}
\label{app:mad-attack}
The MAD attack follows the observation-perturbation principle of maximizing the action deviation induced by the perturbed observation. Let
\begin{equation}
    a^{0}_{i,t}
    =
    \pi_i(s_{i,t})
\end{equation}
be the nominal action of compromised agent $i$. MAD chooses perturbations that maximize the distance between the nominal and perturbed actions:
\begin{equation}
    \mathcal{L}_{\mathrm{MAD}}
    =
    \sum_{i\in\mathcal{M}}
    \left\|
        \pi_i(s_{i,t}+\delta_{i,t})
        -
        a^{0}_{i,t}
    \right\|_2^2 .
\end{equation}
The perturbation is obtained by projected gradient ascent:
\begin{equation}
    \delta^{k+1}_{i,t}
    =
    \Pi_{\|\delta\|_p\le\epsilon}
    \left(
        \delta^{k}_{i,t}
        +
        \eta
        \nabla_{\delta_{i,t}}
        \mathcal{L}_{\mathrm{MAD}}
    \right).
\end{equation}
MAD is reward- and cost-agnostic. It can strongly disturb the victim policy, but it does not explicitly target safety cost increase or reward preservation.

\begin{algorithm}[ht]
\caption{Maximal Action Difference (MAD) Attack}
\label{alg:mad-attack}
\begin{algorithmic}[1]
\Require Local policy inputs $\{s_{i,t}\}_{i\in\mathcal{M}}$, policies $\{\pi_i\}_{i\in\mathcal{M}}$, Perturbation Budget $\epsilon$.
\Require Attack iterations $K$, step size $\eta$.
\Ensure Adversarial observations $\{\hat{s}_{i,t}\}_{i\in\mathcal{M}}$.

\For{each compromised agent $i\in\mathcal{M}$}
    \State Compute nominal action:
    \[
    a^0_{i,t}\leftarrow \pi_i(s_{i,t}).
    \]
    \State Initialize perturbation $\delta^{(0)}_{i,t}\leftarrow 0$.
\EndFor

\For{$k=0,\ldots,K-1$}
    \State Define the MAD objective:
    \[
    \mathcal{L}_{\mathrm{MAD}}
    =
    \sum_{i\in\mathcal{M}}
    \left\|
    \pi_i(s_{i,t}+\delta^{(k)}_{i,t})
    -
    a^0_{i,t}
    \right\|_2^2 .
    \]
    \For{each compromised agent $i\in\mathcal{M}$}
        \State Update perturbation by projected gradient ascent:
        \[
        \delta^{(k+1)}_{i,t}
        \leftarrow
        \Pi_{\|\delta\|_p\le\epsilon}
        \left[
        \delta^{(k)}_{i,t}
        +
        \eta
        \nabla_{\delta_{i,t}}
        \mathcal{L}_{\mathrm{MAD}}
        \right].
        \]
    \EndFor
\EndFor

\For{each compromised agent $i\in\mathcal{M}$}
    \State Set $\hat{s}_{i,t}\leftarrow s_{i,t}+\delta^{(K)}_{i,t}$.
\EndFor

\State \Return $\{\hat{s}_{i,t}\}_{i\in\mathcal{M}}$.
\end{algorithmic}
\end{algorithm}

\subsection{Max-Cost Attack}
\label{app:mc-attack}

The MC attack is a critic-guided safe RL attack that directly optimizes the observation perturbations toward higher predicted safety cost. As shown in Algorithm~\ref{alg:mc-mr-attack}, the attack starts by initializing the perturbation $\delta^{(0)}_{i,t}$ as zero for each compromised agent $i\in\mathcal{M}$. At each attack iteration, the adversary first constructs perturbed local policy inputs for compromised agents, while keeping the inputs of benign agents unchanged. These perturbed inputs are then passed through the victim policies $\boldsymbol{\pi}$ to obtain the induced joint action $\widetilde{a}_t$.

Let $Q_C(s_t,a_t)$ denote the centralized cost critic of the victim model. For WCMASAC, $Q_C$ is replaced by the corresponding risk-sensitive estimate derived from the distributional cost critic. MC uses the attack objective
\begin{equation}
    \mathcal{L}_{\mathrm{MC}}
    =
    Q_C
    \left(
        s_t,
        \widetilde{a}_t
    \right),
\end{equation}
where $\widetilde{a}_t$ is induced by the perturbed observations of compromised agents. The adversary then updates the perturbation by ascending the gradient of the cost critic:
\begin{equation}
    \delta^{k+1}_{i,t}
    =
    \Pi_{\|\delta\|_p\le\epsilon}
    \left(
        \delta^{k}_{i,t}
        +
        \eta
        \nabla_{\delta_{i,t}}
        Q_C(s_t,\widetilde{a}_t)
    \right).
\end{equation}
The projection operator ensures that the updated perturbation remains within the Perturbation Budget $\epsilon$. Therefore, MC explicitly searches for observation perturbations that make the compromised agents induce a joint action with higher predicted safety cost.

MC is more safety-targeted than Random perturbation and MAD because it directly uses the cost critic to guide the attack direction. However, its objective only maximizes the predicted safety cost and does not consider whether the resulting behavior preserves task reward.  

\begin{algorithm}[ht]
\caption{Critic-Guided Attack for MC and MR}
\label{alg:mc-mr-attack}
\begin{algorithmic}[1]
\Require Global state $s_t$, local policy inputs $\{s_{i,t}\}_{i\in\mathcal{N}}$, policies $\boldsymbol{\pi}$, critic $Q_\chi$, compromised set $\mathcal{M}$.
\Require Attack type $\chi\in\{C,R\}$, Perturbation Budget $\epsilon$, attack iterations $K$, step size $\eta$.
\Ensure Adversarial observations $\{\hat{s}_{i,t}\}_{i\in\mathcal{M}}$.

\For{each compromised agent $i\in\mathcal{M}$}
    \State Initialize perturbation $\delta^{(0)}_{i,t}\leftarrow 0$.
\EndFor

\For{$k=0,\ldots,K-1$}
    \State Construct perturbed local policy inputs:
    \[
    \tilde{s}^{(k)}_{i,t}
    =
    \begin{cases}
    s_{i,t}+\delta^{(k)}_{i,t}, & i\in\mathcal{M},\\
    s_{i,t}, & i\notin\mathcal{M}.
    \end{cases}
    \]

    \State Compute induced joint action:
    \[
    \widetilde{\boldsymbol{a}}^{(k)}_t
    \leftarrow
    \boldsymbol{\pi}(\tilde{\boldsymbol{s}}^{(k)}_t).
    \]

    \State Define the critic-guided objective:
    \[
    \mathcal{L}_{\chi}
    =
    Q_\chi(s_t,\widetilde{\boldsymbol{a}}^{(k)}_t),
    \qquad
    \chi\in\{C,R\}.
    \]
    \Statex \hspace{\algorithmicindent} \textit{For MC, $Q_\chi=Q_C$ and $\mathcal{L}_{\chi}=\mathcal{L}_{\mathrm{MC}}$.}
    \Statex \hspace{\algorithmicindent} \textit{For MR, $Q_\chi=Q_R$ and $\mathcal{L}_{\chi}=\mathcal{L}_{\mathrm{MR}}$.}

    \For{each compromised agent $i\in\mathcal{M}$}
        \State Update perturbation by projected gradient ascent:
        \[
        \delta^{(k+1)}_{i,t}
        \leftarrow
        \Pi_{\|\delta\|_p\le\epsilon}
        \left[
        \delta^{(k)}_{i,t}
        +
        \eta
        \nabla_{\delta_{i,t}}
        \mathcal{L}_{\chi}
        \right].
        \]
    \EndFor
\EndFor

\For{each compromised agent $i\in\mathcal{M}$}
    \State Set $\hat{s}_{i,t}\leftarrow s_{i,t}+\delta^{(K)}_{i,t}$.
\EndFor

\State \Return $\{\hat{s}_{i,t}\}_{i\in\mathcal{M}}$.
\end{algorithmic}
\end{algorithm}

\subsection{Max-Reward Attack}
\label{app:mr-attack}

The MR attack follows a similar projected-gradient procedure to MC, but it uses the victim reward critic instead of the cost critic. As shown in Algorithm~\ref{alg:mc-mr-attack}, the attack first initializes the perturbation $\delta^{(0)}_{i,t}$ as zero for each compromised agent $i\in\mathcal{M}$. At each iteration, the adversary perturbs only the local policy inputs of compromised agents and keeps the benign agents' inputs unchanged. The perturbed inputs are then fed into the victim policies $\boldsymbol{\pi}$ to produce the induced joint action $\widetilde{a}_t$.

Let $Q_R(s_t,a_t)$ denote the centralized reward critic. MR uses the attack objective
\begin{equation}
    \mathcal{L}_{\mathrm{MR}}
    =
    Q_R
    \left(
        s_t,
        \widetilde{a}_t
    \right).
\end{equation}
The perturbation is updated by projected gradient ascent:
\begin{equation}
    \delta^{k+1}_{i,t}
    =
    \Pi_{\|\delta\|_p\le\epsilon}
    \left(
        \delta^{k}_{i,t}
        +
        \eta
        \nabla_{\delta_{i,t}}
        Q_R(s_t,\widetilde{a}_t)
    \right).
\end{equation}
The projection step enforces the Perturbation Budget $\epsilon$, ensuring that the adversarial observation remains bounded. Different from MC, MR does not directly maximize the predicted safety cost. Instead, it searches for perturbations that lead the victim policies toward actions with higher predicted task reward.

Although MR is not explicitly safety-targeted, it can still induce unsafe behavior in safe RL. In some environments, high-reward actions may lie close to safety boundaries or require more aggressive behavior, which can increase safety cost while maintaining nominal task reward. Therefore, MR serves as a baseline for evaluating stealthiness.
It helps assess whether an attack can preserve task reward while increasing safety cost.

\section{Supplementary Experiment Results}
\subsection{Supplementary Results for Section 5.2}
This section provides additional results on how the Perturbation Budget affects attack performance on the ShadowHandReOrientation task for the remaining victim algorithms. We report the results for the other three algorithms below.

\begin{figure}[ht]
  \small
  \centering
  \makebox[\linewidth][c]{%
  \includegraphics[width=1.1\linewidth,height=4.1cm]{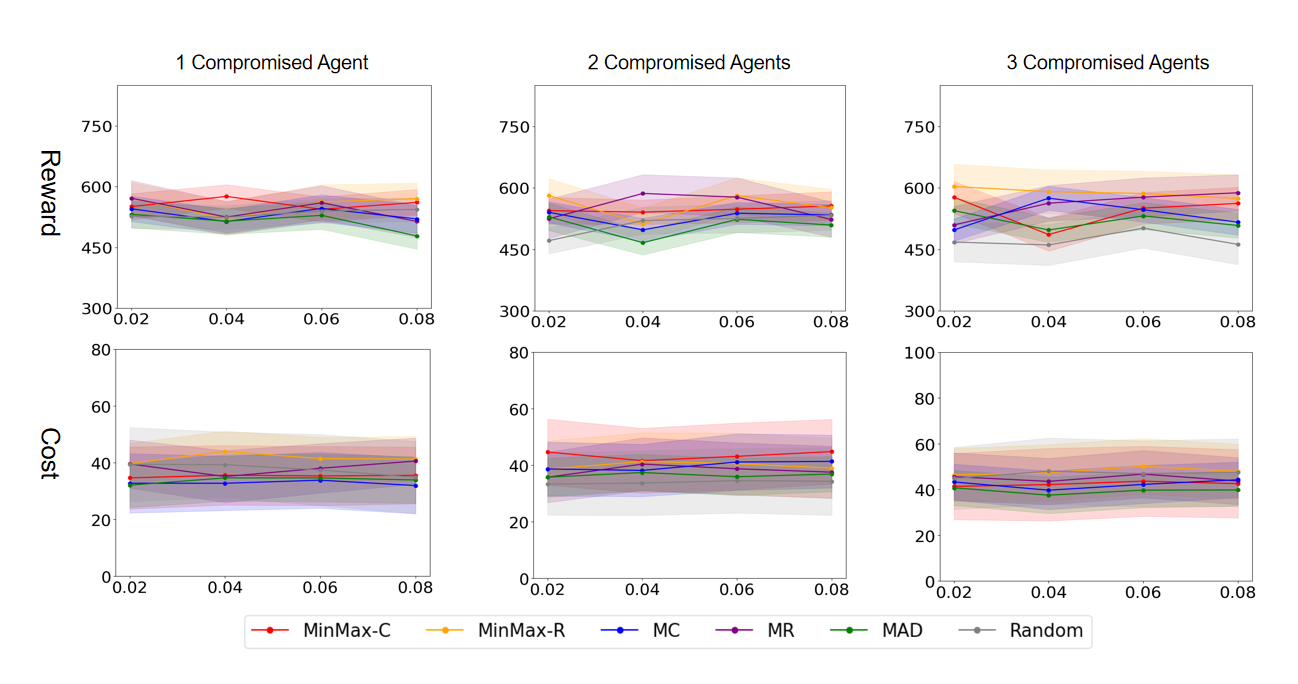}%
}
  \caption{Attack performance on HASAC-Lag under different numbers of compromised agents and Perturbation Budgets. The x-axis denotes the Perturbation Budget $\epsilon$.}
  \label{fig:app1}
\end{figure}

Figure~\ref{fig:app1} provides additional results on HASAC-Lag under different Perturbation Budgets $\epsilon$ and Compromised-Agent Budgets $m$. Across all settings, the task reward remains relatively stable, indicating that the attacks do not cause obvious reward collapse. In contrast, the cumulative cost varies more clearly across attack methods.

The MinMax variants consistently achieve the strongest or near-strongest safety attacks. With one compromised agent, MinMax-R induces the highest cost across most perturbation budgets while maintaining nominal task reward. With two compromised agents, MinMax-C produces the largest cost, and its cost increases mildly as $\epsilon$ grows. With three compromised agents, MinMax-R again achieves the highest cost, while MinMax-C remains competitive. Overall, the results support the same trend observed in the main experiments: critic-guided MinMax variants induce higher safety cost than non-structured baselines while preserving task reward, demonstrating stronger stealthy safety attack performance on HASAC-Lag.

\begin{figure}[ht]
  \small
  \centering
  \makebox[\linewidth][c]{%
  \includegraphics[width=1.1\linewidth,height=4.1ccm]{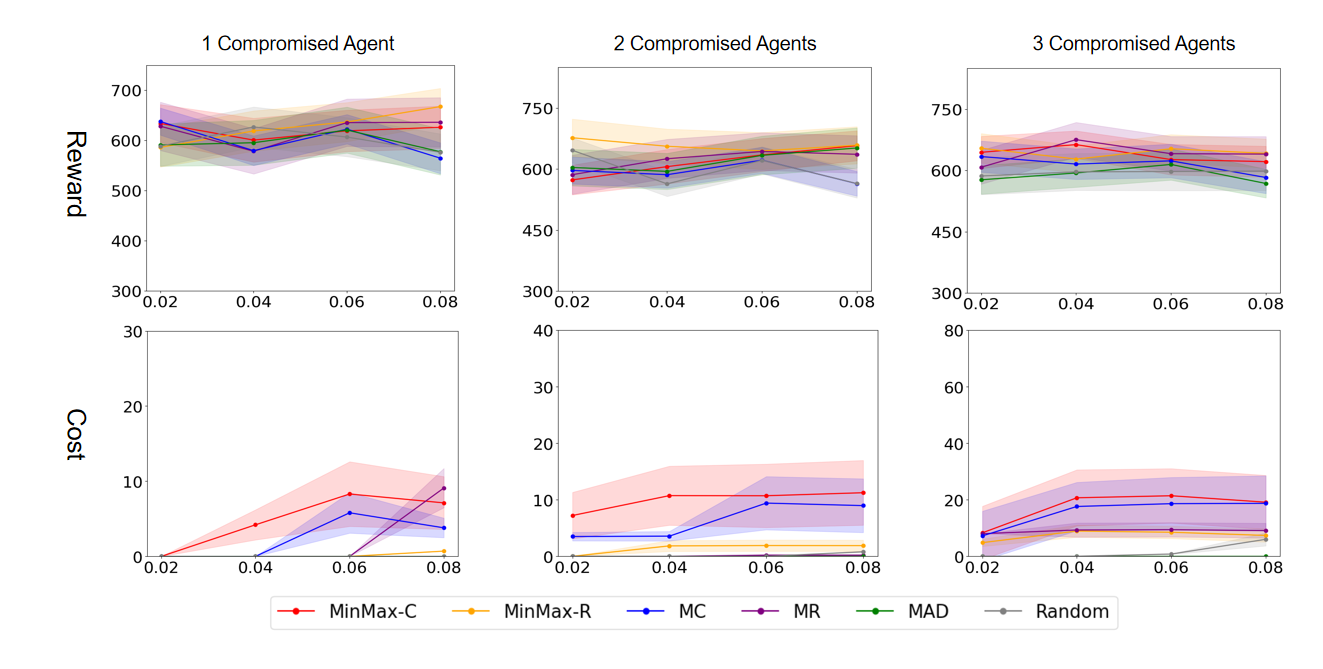}%
}
  \caption{Attack performance on MAPPO-Lag under different numbers of compromised agents and Perturbation Budgets. The x-axis denotes the Perturbation Budget $\epsilon$.}
  \label{fig:app2}
\end{figure}

Figure~\ref{fig:app2} reports additional results on MAPPO-Lag under different Perturbation Budgets $\epsilon$ and Compromised-Agent Budgets $m$. The reward curves do not exhibit a clear monotonic trend as $\epsilon$ increases, and the relatively large standard deviations indicate substantial variability across runs. 

The cost curves show a clearer separation among attack methods. MinMax-C consistently induces higher safety cost than non-structured baselines, especially when two or three agents are compromised. MC also increases cost in several settings, but MinMax-C generally achieves a stronger cost increase under comparable reward levels. In contrast, Random, MAD, and MinMax-R remain close to zero cost in most settings, suggesting limited safety-violation capability on MAPPO-Lag. Overall, the results indicate that the cost-critic-driven MinMax-C variant is the most effective attack on MAPPO-Lag.

\begin{figure}[ht]
  \small
  \centering
  \makebox[\linewidth][c]{%
  \includegraphics[width=1.1\linewidth,height=4.1cm]{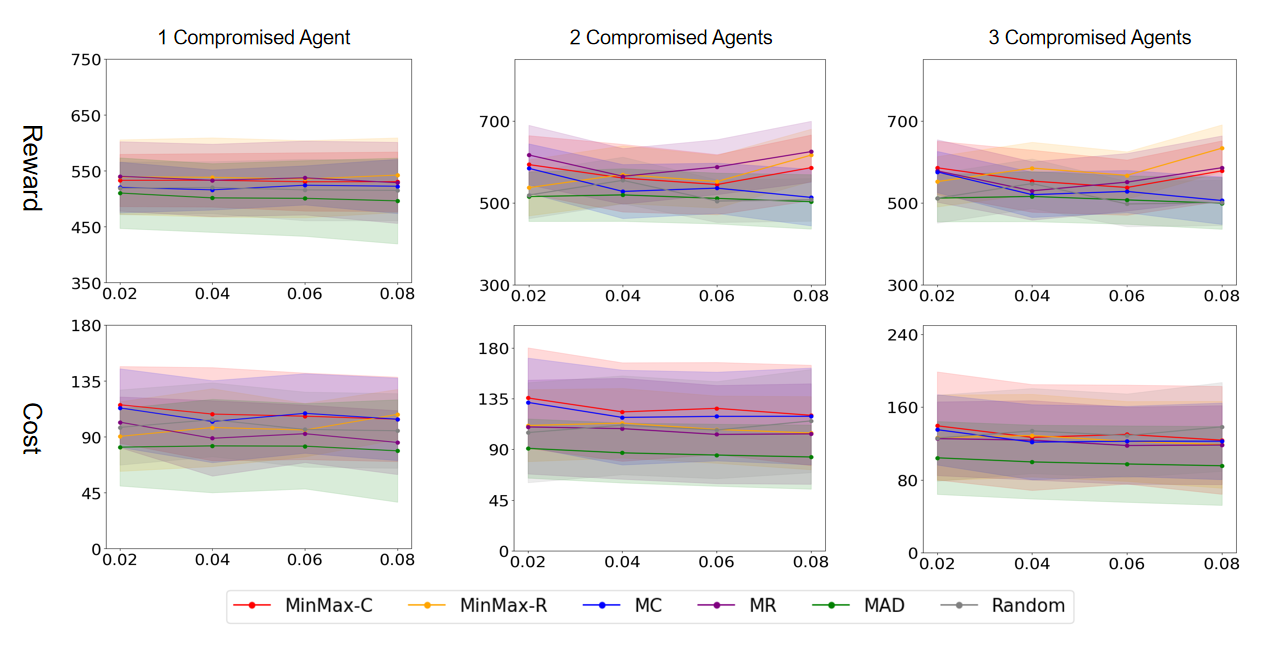}%
}
  \caption{Attack performance on MACPO under different numbers of compromised agents and Perturbation Budgets. The x-axis denotes the Perturbation Budget $\epsilon$.}
  \label{fig:app3}
\end{figure}

Figure~\ref{fig:app3} reports additional results on MACPO under different Perturbation Budgets $\epsilon$ and Compromised-Agent Budgets $m$. Similar to MAPPO-Lag, the reward curves do not show a clear monotonic trend as $\epsilon$ increases, and the shaded regions indicate relatively large variability across runs.

For safety cost, the separation across methods is also not strictly monotonic with respect to $\epsilon$. Nevertheless, critic-guided methods generally induce higher costs than non-structured baselines. MinMax-C achieves competitive or leading cost levels in most settings, especially when one or two agents are compromised. Overall, these results further support that MinMax Attack can increase safety cost without causing consistent reward collapse, although the effect of increasing $\epsilon$ is environment- and algorithm-dependent rather than strictly monotonic.

\subsection{Supplementary Results for Section 5.3}
This section provides additional results on how the Perturbation Budget affects attack performance on the Safe Multi-Agent MuJoCo HalfCheetah task. We report results for two representative victim algorithms below.

\begin{figure}[ht]
  \small
  \centering
  \makebox[\linewidth][c]{%
  \includegraphics[width=1.1\linewidth,height=3.8cm]{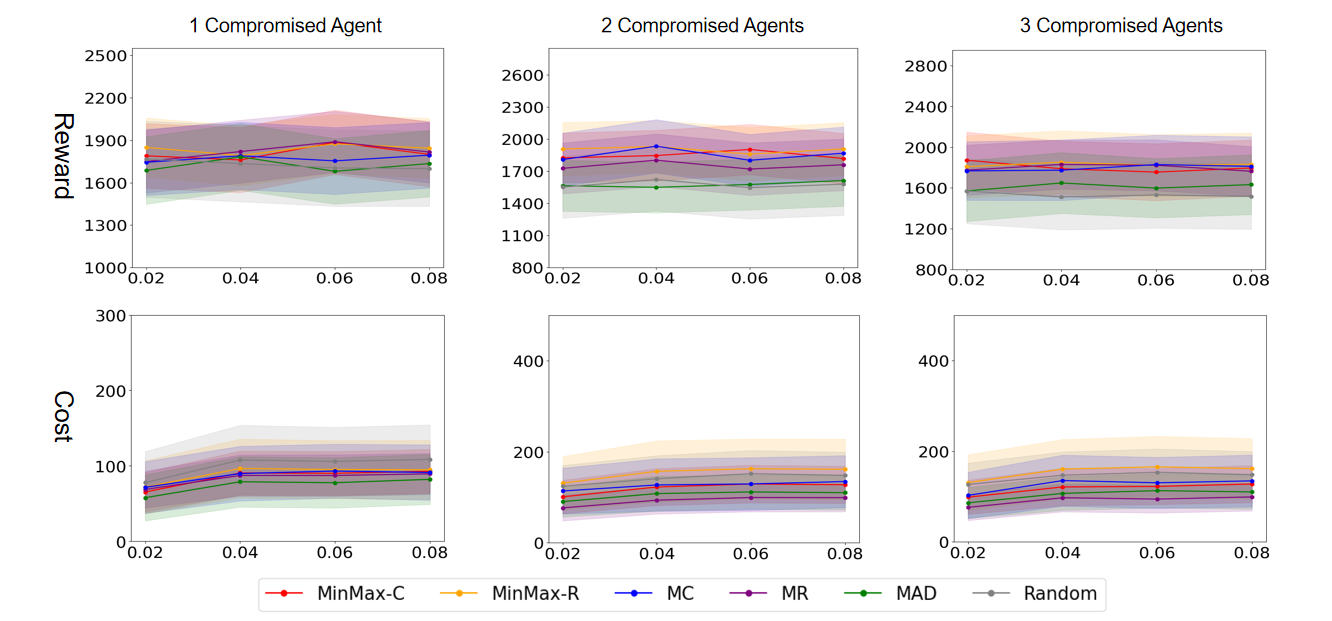}%
}
  \caption{Attack performance on MACPO under different numbers of compromised agents and Perturbation Budgets. The x-axis denotes the Perturbation Budget $\epsilon$.}
  \label{fig:app4}
\end{figure}

Figure~\ref{fig:app4} reports the attack performance on MACPO under different Perturbation Budgets $\epsilon$ and Compromised-Agent Budgets $m$. The reward curves remain relatively stable across perturbation budgets. The effect of increasing $\epsilon$ is also not strictly monotonic, especially after $\epsilon=0.04$, where both reward and cost tend to fluctuate within a similar range.

For safety cost, increasing the Compromised-Agent Budget generally leads to a higher cost level than increasing the Perturbation Budget alone. MinMax-R and MC induce relatively high costs across most settings, while MinMax-C remains competitive. 

\begin{figure}[ht]
  \small
  \centering
  \makebox[\linewidth][c]{%
  \includegraphics[width=1.1\linewidth,height=4.1cm]{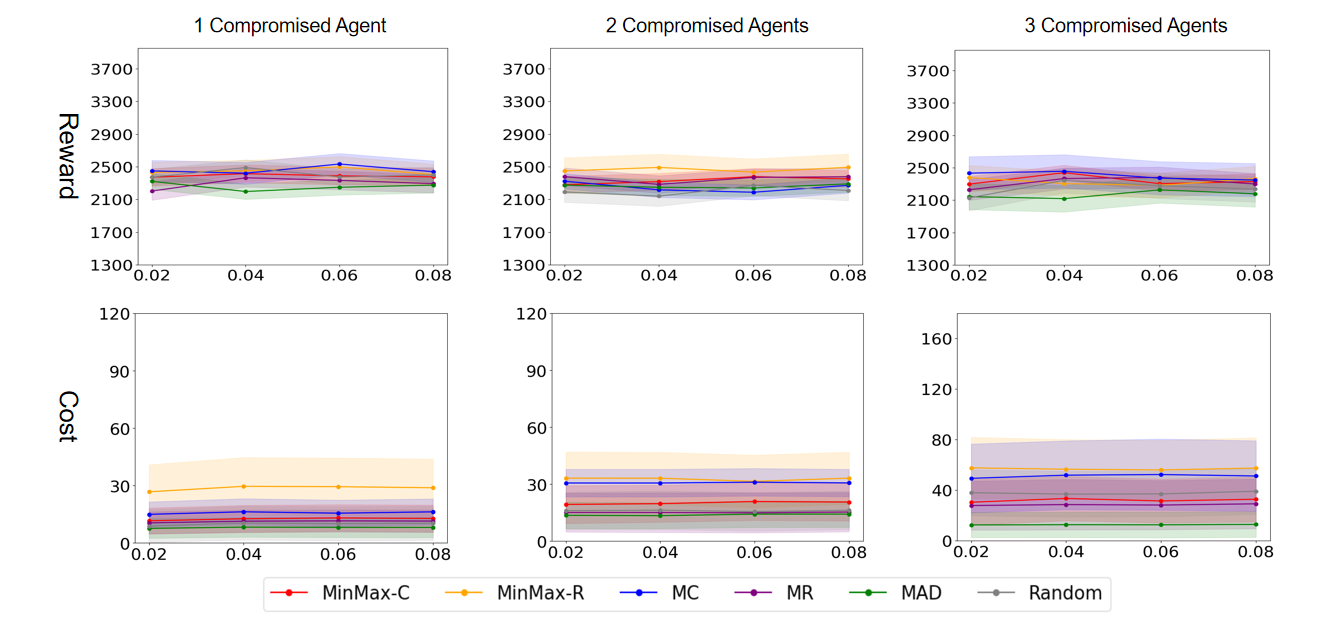}%
}
  \caption{Attack performance on MAPPO-Lag under different numbers of compromised agents and Perturbation Budgets. The x-axis denotes the Perturbation Budget $\epsilon$.}
  \label{fig:app5}
\end{figure}

Figure~\ref{fig:app5} reports additional results under different Perturbation Budgets $\epsilon$ and Compromised-Agent Budgets $m$ on MAPPO-Lag in other environment. The reward curves remain relatively stable across different $\epsilon$ values, and no clear monotonic trend is observed.  
The cost curves also show limited sensitivity to $\epsilon$ in most settings. Instead, the Compromised-Agent Budget $m$ has a more visible effect on safety cost, especially when moving from one to three compromised agents. MinMax-R generally induces higher costs than non-structured baselines, while maintaining comparable rewards. Overall, these results indicate that the attack effect is more strongly influenced by the number of compromised agents than by small changes in the Perturbation Budget.

\section{Additional Experimental Analyses}
\label{app:additional-experiments}
This section supplements the main evaluation with paired statistical comparisons. Table~\ref{tab:app-significance} reports comparisons between MinMax-C and MC, and between MinMax-R and MR. The quantities $\Delta R$ and $\Delta C$ denote the mean paired differences in episodic reward and safety cost, respectively, computed as MinMax minus the corresponding baseline and reported with $95\%$ confidence intervals. For reward, we conduct a one-sided non-inferiority test with a pre-specified margin of $5\%$ of the corresponding baseline reward mean, testing $H_0:\Delta R\le-\Delta_{\mathrm{NI}}$ against $H_1:\Delta R>-\Delta_{\mathrm{NI}}$, where $\Delta_{\mathrm{NI}}=0.05|R_{\mathrm{base}}|$. For safety cost, we conduct a one-sided superiority test, testing $H_0:\Delta C\le0$ against $H_1:\Delta C>0$. The corresponding $p$-values are denoted by $p_{\mathrm{NI}}^R$ and $p_{\mathrm{sup}}^C$, respectively. Shaded bold entries indicate reported $p<0.05$.

\begin{table*}[ht]
\centering
\setlength{\tabcolsep}{1.0pt}         
\renewcommand{\arraystretch}{1.5}   %
\makebox[\textwidth][c]{%
\resizebox{\textwidth}{!}{%

\begin{tabular}{ll|cccc|cccc}
\toprule
Victim Model & $m$ &
 
\multicolumn{4}{c|}{\textbf{MinMax-C vs. MC}}&
\multicolumn{4}{c}{\textbf{MinMax-R vs. MR}}\\
 
\cmidrule(lr){3-6}
\cmidrule(lr){7- 10}
 
& & $\Delta R$ [95\% CI] &$p_{\mathrm{NI}}^{R}$&$\Delta C$ [95\% CI]&$p_{\mathrm{sup}}^{C}$  
& $\Delta R$ [95\% CI] &$p_{\mathrm{NI}}^{R}$&$\Delta C$ [95\% CI]&$p_{\mathrm{sup}}^{C}$   \\
\midrule

\multirow{3}{*}{\shortstack{MACPO}}
 
& 1  &$6.20\,[-17.28,29.68]$ & \cellcolor{gray!20}\textbf{$0.009$} & $9.60\,[2.36,16.84]$ & \cellcolor{gray!20}\textbf{$0.011$}     
    &$6.60\,[-12.79,25.99]$ & \cellcolor{gray!20}\textbf{$0.005$} & $6.20\,[4.71,7.69]$ & \cellcolor{gray!20}\textbf{$<0.001$}
    
    \\
& 2 
    &$8.80\,[-30.58,48.18]$ & \cellcolor{gray!20}\textbf{$0.033$} & $7.10\,[-1.23,15.43]$ & \cellcolor{gray!20}\textbf{$0.039$}
    &$7.20\,[-36.09,50.49]$ & \cellcolor{gray!20}\textbf{$0.046$} & $7.30\,[0.84,13.76]$ & \cellcolor{gray!20}\textbf{$0.017$}
    \\
& 3 
 
    &$9.60\,[-28.82,48.02]$ & \cellcolor{gray!20}\textbf{$0.030$} & $7.20\,[3.60,10.80]$ & \cellcolor{gray!20}\textbf{$0.003$}
    &$16.50\,[-39.05,72.05]$ & \cellcolor{gray!20}\textbf{$0.046$} & $4.60\,[2.74,6.46]$ & \cellcolor{gray!20}\textbf{$0.001$}\\
 
    \midrule
\multirow{3}{*}{\shortstack{MAPPO-Lag}}
& 1 
    &$-3.40\,[-18.42,11.62]$ & \cellcolor{gray!20}\textbf{$0.003$} & $2.90\,[-0.09,5.89]$ & \cellcolor{gray!20}\textbf{$0.027$}
    &
    $0.70\,[-35.36,36.76]$ & \cellcolor{gray!20}\textbf{$0.033$} & $0.50\,[0.25,0.75]$ & \cellcolor{gray!20}\textbf{$0.003$}\\
& 2 
 
    &$12.90\,[-21.79,47.59]$ & \cellcolor{gray!20}\textbf{$0.012$} & $1.30\,[0.09,2.51]$ & \cellcolor{gray!20}\textbf{$0.020$}
    &$1.30\,[-4.64,7.24]$ & \cellcolor{gray!20}\textbf{$<0.001$} & $1.70\,[0.39,3.01]$ & \cellcolor{gray!20}\textbf{$0.011$}\\
& 3 
 
    &$3.10\,[-25.09,31.29]$ & \cellcolor{gray!20}\textbf{$0.014$} & $2.80\,[0.03,5.57]$ & \cellcolor{gray!20}\textbf{$0.024$}
    &$12.10\,[-34.57,58.77]$ & \cellcolor{gray!20}\textbf{$0.029$} & $3.50\,[1.38,5.62]$ & \cellcolor{gray!20}\textbf{$0.005$}\\
  \midrule

\multirow{3}{*}{\shortstack{WCMASAC}}
& 1 
 
    &$5.90\,[-17.66,29.46]$ & \cellcolor{gray!20}\textbf{$0.006$} & $3.30\,[0.03,6.57]$ & \cellcolor{gray!20}\textbf{$0.024$}
    &$6.10\,[-39.55,51.75]$ & \cellcolor{gray!20}\textbf{$0.040$} & $1.40\,[-0.22,3.02]$ & \cellcolor{gray!20}\textbf{$0.037$}\\
& 2 
  
    &$5.20\,[-28.38,38.78]$ & \cellcolor{gray!20}\textbf{$0.019$} & $2.70\,[2.20,3.20]$ & \cellcolor{gray!20}\textbf{$<0.001$}
    &$9.60\,[-26.35,45.55]$ & \cellcolor{gray!20}\textbf{$0.016$} & $7.20\,[-1.39,15.79]$ & \cellcolor{gray!20}\textbf{$0.040$}\\
& 3
 
    &$4.90\,[-26.90,36.70]$ & \cellcolor{gray!20}\textbf{$0.017$} & $5.50\,[-1.57,12.57]$ & \cellcolor{gray!20}\textbf{$0.048$}
    &$8.30\,[-40.63,57.23]$ & \cellcolor{gray!20}\textbf{$0.041$} & $7.80\,[7.43,8.17]$ & \cellcolor{gray!20}\textbf{$<0.001$}\\
  \midrule
 
\multirow{3}{*}{\shortstack{HASAC-Lag}}
& 1 
 
    &$-2.30\,[-5.78,1.18]$ & \cellcolor{gray!20}\textbf{$<0.001$} & $1.50\,[0.21,2.79]$ & \cellcolor{gray!20}\textbf{$0.016$}
    & $1.90\,[-17.27,21.07]$ & \cellcolor{gray!20}\textbf{$0.006$} & $3.40\,[1.26,5.54]$ & \cellcolor{gray!20}\textbf{$0.006$}\\
& 2 
  
    &$9.80\,[-14.11,33.71]$ & \cellcolor{gray!20}\textbf{$0.007$} & $2.00\,[-0.59,4.59]$ & \cellcolor{gray!20}\textbf{$0.049$}
    &$3.90\,[-20.54,28.34]$ & \cellcolor{gray!20}\textbf{$0.010$} & $1.80\,[-0.19,3.79]$ & \cellcolor{gray!20}\textbf{$0.033$}\\
& 3 
 
    &$5.60\,[-37.19,48.39]$ & \cellcolor{gray!20}\textbf{$0.050$} & $4.60\,[0.63,8.57]$ & \cellcolor{gray!20}\textbf{$0.016$}
    &$12.10\,[-21.30,45.50]$ & \cellcolor{gray!20}\textbf{$0.014$} & $3.40\,[1.02,5.78]$ & \cellcolor{gray!20}\textbf{$0.008$}\\
\midrule
\end{tabular}
}}
\caption{Statistical comparison of MinMax-C vs.\ MC and MinMax-R vs.\ MR, reporting reward non-inferiority and safety cost superiority with $95\%$ confidence intervals. Differences are MinMax minus the corresponding baseline; $m$ is the number of compromised agents. Shaded bold entries have reported one-sided $p<0.05$.}
\label{tab:app-significance}
\end{table*}

\end{document}